\documentclass[lettersize,journal]{IEEEtran}
\usepackage{amsmath,amsfonts}
\usepackage{algorithmic}
\usepackage{algorithm}
\usepackage{array}
\usepackage[caption=false,font=normalsize,labelfont=sf,textfont=sf]{subfig}
\usepackage{textcomp}
\usepackage{stfloats}
\usepackage{url}
\usepackage{verbatim}
\usepackage{graphicx}
\usepackage{cite}
\usepackage{svg}
\usepackage{caption}
\usepackage{colortbl}
\usepackage{multirow}
\usepackage{booktabs} 
\usepackage{marginnote}
\usepackage{hhline}
\usepackage{multirow}
\usepackage{amssymb} 
\begin{document}

\title{DCP-Net: A Distributed Collaborative Perception Network for Remote Sensing Semantic Segmentation}

\author{Zhechao~Wang, 
	Peirui~Cheng, 
	Shujing~Duan, 
	Kaiqiang~Chen,
	Zhirui~Wang, \IEEEmembership{Member,~IEEE},
	Xinming Li, 
	Xian~Sun, \IEEEmembership{Senior Member,~IEEE}
	
	\thanks{This work was supported by the National Nature Science Foundation of China under Grant 62171436, Grant 62076241, and Grant 62201550.
		\textit{(Corresponding author: Zhirui Wang.)}}
	\thanks{Zhechao Wang, Shujing Duan and Xian Sun are with the Aerospace Information Research Institute, Chinese Academy of Sciences, Beijing 100190, China, also with the Key Laboratory of Network Information System Technology (NIST), Aerospace Information Research Institute, Chinese Academy of Sciences, Beijing 100190, China, also with the University of Chinese Academy of Sciences, Beijing 100190, China, and also with the School of Electronic, Electrical and Communication Engineering, University of Chinese Academy of Sciences, Beijing 100190, China (e-mail: wangzhechao21@mails.ucas.ac.cn;  duanshujing21@mails.ucas.ac.cn; sunxian@aircas.ac.cn).}
	\thanks{Peirui Cheng, Kaiqiang Chen, Zhirui Wang and Xinming Li are with the Aerospace Information Research Institute, Chinese Academy of Sciences, Beijing 100094, China, and also with the Key Laboratory of Network Information System Technology (NIST), Aerospace Information Research Institute, Chinese Academy of Sciences, Beijing 100190, China (e-mail: chengpr@aircas.ac.cn;
		chenkq@aircas.ac.cn;
		zhirui1990@126.com;  13911729321@139.com).}
}

\markboth{Journal of \LaTeX\ Class Files,~Vol.~14, No.~8, August~2021}%
{Shell \MakeLowercase{\textit{et al.}}: A Sample Article Using IEEEtran.cls for IEEE Journals}


\maketitle

\begin{abstract}
Onboard intelligent processing is widely applied in emergency tasks in the field of remote sensing. However, it is predominantly confined to an individual platform with a limited observation range as well as susceptibility to interference, resulting in limited accuracy. 
Considering the current state of multi-platform collaborative observation, this article innovatively presents a distributed collaborative perception network called DCP-Net. Firstly, the proposed DCP-Net helps members to enhance perception performance by integrating features from other platforms. Secondly, a self-mutual information match module is proposed to identify collaboration opportunities and select suitable partners, prioritizing critical collaborative features and reducing redundant transmission cost.
Thirdly, a related feature fusion module is designed to address the misalignment between local and collaborative features, improving the quality of fused features for the downstream task.
We conduct extensive experiments and visualization analyses using three semantic segmentation datasets, including Potsdam, iSAID and DFC23. The results demonstrate that DCP-Net outperforms the existing methods comprehensively, improving mIoU by 2.61\% $ \sim $ 16.89\% at the highest collaboration efficiency, which promotes the performance to a state-of-the-art level.
\end{abstract}


\section{Introduction}
Intelligent remote sensing platforms, such as satellites and drones, have demonstrated exceptional performance in onboard real-time processing tasks, which has been applied in disaster detection \cite{wang2023hierarchical}, maritime monitoring \cite{xu2022lmo, chen2022contrastive}, water-body extraction \cite{parajuli2022attentional} and urban mapping \cite{gu2022intensity, gu2023spatial}.
With the widespread deployment of intelligent remote sensing platforms, these platforms possess the capability of collaborative observation \cite{qin2022new, li2022orbit} and swarm internal communication \cite{zhang2022progress, campion2018uav}, laying the foundation for collaborative perception.

\begin{figure}
	\centering
	\includegraphics[width=0.45\textwidth]{"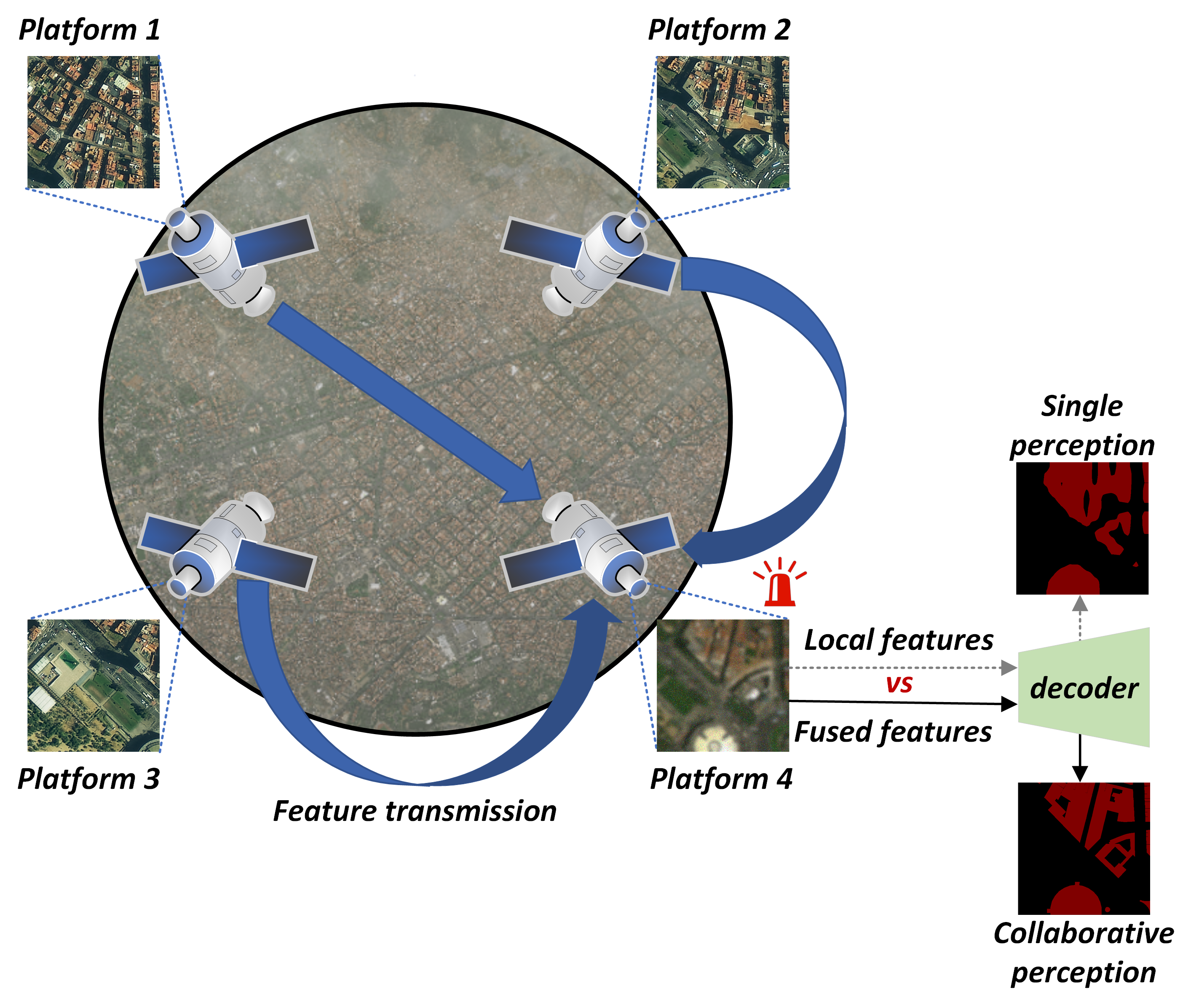"}
	\caption{\textbf{Illustration of multiple remote sensing platforms' collaborative perception.} The members of the collaborative group can enhance their local perception with the help of collaborative information from other platforms.} 
	\label{figure.1}
\end{figure}

Collaborative perception in remote sensing refers to the information sharing among multiple intelligent platforms to enhance the perceptual ability, as depicted in Fig.\ref{figure.1}.
Multiple intelligent remote sensing platforms form a collaborative group to jointly observe an identical scene from different angles.
Within this group, platforms can share their local information to expand the scope of observation and leverage the collaborative information to compensate for the limitations of individual platform perception.
Given the numerous benefits of collaborative perception, it is crucial to further develop this technology in the field of remote sensing.

However, there exist several significant challenges that need to be addressed while implementing collaborative perception in remote sensing. 
Firstly, high-speed mobility \cite{wang2013large, liu2011echo} and long-distance transmission \cite{li2016wireless} contribute to limited bandwidth \cite{juan2012enhanced, ma2022high} and weak communication \cite{warnick2018phased} among platforms, posing obstacles to the interaction in collaborative perception. Consequently, this can lead to performance degradation and even complete interaction failure. Although some collaborative methods alleviate this issue by reducing the amount of transmitted data through the utilization of processed features rather than raw data \cite{zhang2023dcnnet}, \cite{wang2023dcm}, the frequent transmission of data streams and the indiscriminative fully connected interaction mode remain insufficient for the weak communication in remote sensing.
Secondly, the multi-angle characteristic of remote sensing introduces variations in observations across different platforms for an identical scene, resulting in information misalignment between platforms. Given that targets in remote sensing scenes are typically small and densely distributed, simply adding or concatenating features from different platforms without considering their potential distinctions at the same region can interfere with original observations and deteriorate the perception performance \cite{liu2020who2com,liu2020when2com}.

As to the first issue, we recognize that not all collaboration is necessary, and not all collaborative information is equally crucial. Therefore, we adopt a dynamic on-demand manner rather than relying on frequent fully connected interaction. This approach establishes collaboration only when necessary and prioritizes the transmission of crucial collaboration information, reducing the frequency and volume of transmission while enhancing interaction efficiency during collaborative perception. As a result, it better adapts to the prevalent weak communication environment commonly encountered in remote sensing scenarios.
Regarding the second issue, exploring the correspondence between observations from different perspectives can effectively mitigate the interference caused by feature misalignment on the original local features during collaboration. Consequently, when filtering the collaborative features among different platforms, we establish their correlation with the guidance of the local features. This approach facilitates the extraction of relevant information and ensures the alignment and integration of features, thus guaranteeing the overall performance of collaborative perception.

To implement the solutions mentioned above, this paper introduces the distributed collaborative perception network (DCP-Net) for multiple remote sensing platforms. DCP-Net realizes dynamic interactions through the design of a self-mutual information match (SMIM) module. Based on the local information, inter-platform relevance, and designed strategy, the SMIM module evaluates the necessity of collaboration and identifies suitable platforms for collaboration, effectively avoiding redundant interactions and minimizing the interference caused by low-correlation information. This process contributes to relieving the transmission burden in weak communication while maintaining the perception performance.
Furthermore, this paper presents a related feature fusion (RFF) module to overcome the challenge of information misalignment. The RFF module models the relationship between local and collaborative features and selectively integrates locally required information for feature fusion. This approach effectively mitigates the interference caused by misaligned features. With the implementation of these designed modules, DCP-Net facilitates collaborative perception among remote sensing platforms with low transmission amount and high-quality fusion. 

In conclusion, the main contributions of this article can be summarized as follows:

1) This paper presents an innovative approach called DCP-Net for collaborative remote sensing semantic segmentation.
Our DCP-Net enhances the overall perception performance of platforms through collaboration and achieves state-of-the-art results on Potsdam, iSAID and DFC23.

2) A SMIM module is designed to reduce redundant transmission and avoid irrelevant interference, thereby improving adaptation to the weak communication in remote sensing. This module determines the opportunities and partners for collaboration, achieving a balance between perception enhancement and communication overhead.

3) This paper presents an RFF module to address the
issue of information misalignment and ensure the quality of
the fused feature representation. This module is responsible for feature filtering and facilitates the effective integration of both local and collaborative features.

4) To the best of our knowledge, this is the first paper to investigate collaborative perception through feature interactions among multiple platforms in the field of remote sensing.

\section{Related work}
\subsection{Collaborative Perception}
Collaborative perception is a burgeoning application for multi-agent systems, where agents integrate local observations with those of neighboring agents in a learnable manner to enhance accuracy in perception tasks. This field has gained significant attention, and several works have been established to support and advance research in this area.

Liu et al. \cite{liu2020who2com} are the first to propose the concept of collaborative perception and introduce a multi-stage handshake communication mechanism called Who2com. This mechanism allows neural networks to learn how to compress relevant information for each stage of communication. They also develop a simulated dataset using the AirSim simulator, which is perceived by a group of aerial robots. 
When2com \cite{liu2020when2com}, as an upgraded version of Who2com, reformulates the communication framework by learning how to construct communication groups and determine the optimal timing for communication. The generalizability of When2com is demonstrated in tasks such as multi-agent 3D shape recognition and collaborative semantic segmentation. Hu et al. \cite{hu2022where2comm} introduce Where2com, which aims to optimize communication efficiency by conveying spatially sparse but perceptually essential information. They demonstrate the effectiveness of Where2com in the task of 3D vehicle detection with two modalities of camera and LiDAR. Zhou et al. \cite{zhou2022multi} adopt a general-purpose Graph Neural Network to improve single-robot perception accuracy and resilience to sensor failures and disturbances in multi-robot monocular depth estimation.

In some domains of cross-applications, collaborative perception is increasingly embraced. 
A collaborative perception framework called Swarm-SLAM \cite{lajoie2023swarm} is proposed to tackle the challenge of sharing situational awareness among multiple robots operating in GPS-absent conditions, simultaneously achieving localization and mapping tasks. The primary objective of this framework is to identify inter-robot map links and utilize them to merge the individual local maps, ultimately creating a unified global understanding of the environment.
In the realm of few-shot learning, FS-MAP \cite{fan2023few} is presented as a metric-based framework for air-ground collaboration. The framework incorporates multiple UAVs dedicated to collecting few-shot face samples, along with a self-driving campus delivery vehicle for target queries. It operates by first sending face queries from the vehicle to the UAVs for matching. Next, the UAVs send back similarity scores to the delivery vehicle, which then the vehicle sorts and filters the scores to determine the geographic location of the target for subsequent path planning.
In the field of game AI agent learning, Nash et al. \cite{nash2023herd} introduce a novel paradigm called Herd's Eye View, inspired by the concept of the Bird's Eye View in autonomous vehicles. It employs cooperative perception to enhance the decision-making of RL agents by providing them with global reasoning abilities. In contrast to previous collaborative perception approaches, the method combines reinforcement learning to address both low-level control tasks and high-level planning challenges simultaneously in complex, procedurally generated environments, demonstrating superior performance compared to traditional ego-centric perception models.
While there is limited research on collaborative perception in remote sensing, Gao et al. \cite{gao2023onboard} mention the potential of multi-platform networking and collaboration in modern earth observation systems. This collaboration can lead to acquiring earth observation data with higher accuracy, advanced information dimensions, and higher space-time resolution compared to current systems. With the proliferation of satellite constellations and unmanned aerial vehicle swarms, numerous applications exist for exploring collaborative perception.

\subsection{Remote Sensing Semantic Segmentation}
Semantic segmentation in the field of natural scenes has advanced significantly with deep neural networks.
Long et al. \cite{long2015fully} pioneer the application of deep neural networks to semantic segmentation, replacing fully connected layers with convolution layers and incorporating multi-scale feature fusion techniques. 
U-Net \cite{ronneberger2015u} introduces a decoder that performs feature upsampling and incorporates features from neighboring layers through concatenation. This innovative design aims to effectively recover lost spatial information during the process of semantic segmentation.
Chen et al. \cite{chen2017deeplab} propose the DeepLab series, incorporating dilated convolutions to capture multi-scale contextual information.
In addition to the CNN-based methods mentioned above, there are also novel approaches that leverage transformer-based architectures. 
Zheng et al.\cite{zheng2021rethinking} propose SETR that replaces the traditional CNN encoder with the vision transformer and employs a progressive up-sampling and multi-level feature aggregation for relieving the noise during the overall process. SegFormer \cite{xie2021segformer} utilizes a hybrid architecture that combines transformers with a lightweight multilayer perceptron decoder.

The methods employed for semantic segmentation in remote sensing are typically derived from those utilized in natural scenes and subsequently tailored to specific application scenarios.
Mou et al. \cite{mou2020relation} apply relation-augmented feature representations into FCNs in order to capture long-range spatial relationships among entities in satellite images. 
Diakogiannis et al. \cite{diakogiannis2020resunet} introduce ResUNet, a novel approach that integrates the advantages of ResNet, U-Net, and ASPP architecture to effectively handle objects with diverse sizes in remote sensing imagery.
Niu et al. \cite{niu2021improving} incorporate graph reasoning and disentangled learning to the common decoder architectures to improve the localization and precision of the segmentation results. 
DisOptNet \cite{kang2022disoptnet} adopts knowledge distillation between multi-source aerial images to enhance the spatial consistency and contextual understanding of the segmentation results.
A novel attention-based framework, HMANet \cite{niu2021hybrid} captures global correlations adaptively within the aspects of spaces, channels, and categories more effectively and efficiently.
ST-UNet \cite{he2022swin} demonstrates the great potential of the UNet-like transformer decoder in effectively modeling global information for urban scene segmentation.
These methods mentioned above aim to realize a high-performance semantic segmentation of remote sensing imagery in a centralized data center. For the sake of simplicity and practicality, DCP-Net employs a lightweight FCN-based decoder to facilitate onboard intelligent processing across multiple platforms.

\section{Method}
\subsection{Overview}
\begin{figure*}
	\centering
	\includegraphics[width=1.\textwidth]{"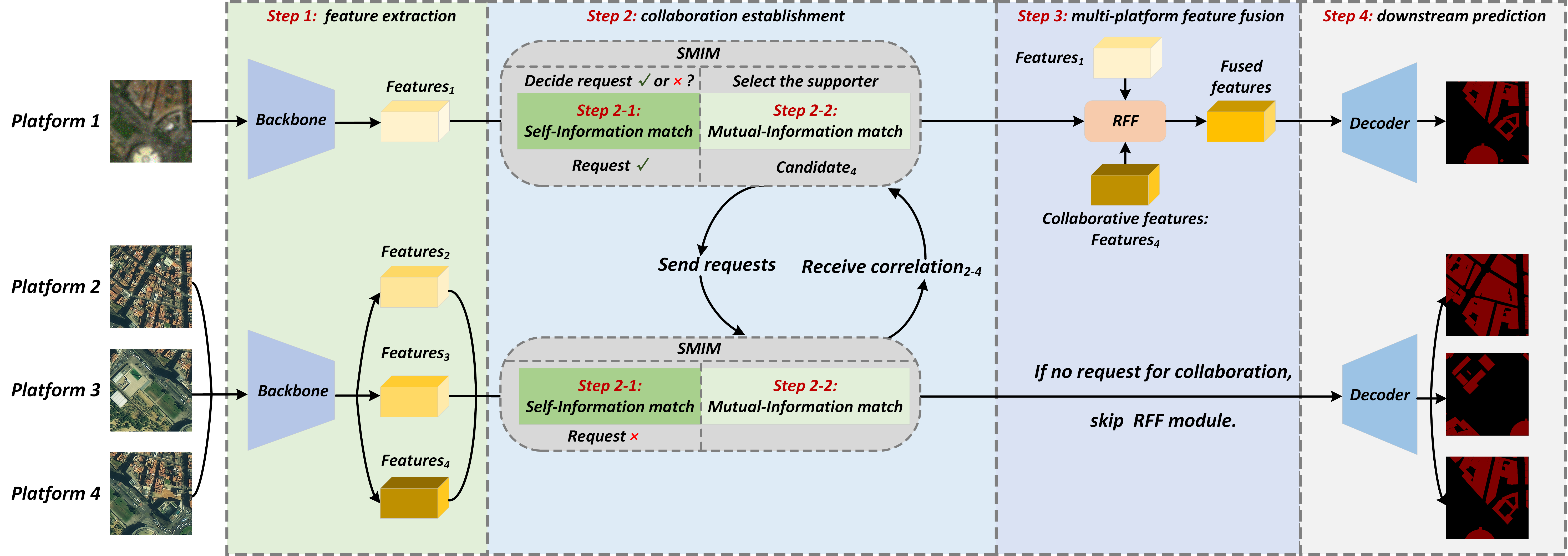"}
	\caption{\textbf{The framework of Distributed Collaborative Perception Network}. Within this framework, the platform that requires collaboration is referred to as a \textbf{requester}. The platforms involved in the collaborative network, excluding the requester, are termed \textbf{candidates}. Among the candidates, the selected platform that acts as a perception helper is referred to as the \textbf{supporter}. In the given scenario, platform 1 acts as the requester, while the remaining platforms serve as candidates for collaboration. Through the SMIM module, platform 4 is chosen as the supporter, responsible for providing collaborative features.
	}
	\label{figure.2}
\end{figure*}

The proposed Distributed Collaborative Perception Network is depicted in Fig.\ref{figure.2}, consisting of four modules, each with its specific role: feature extraction, collaboration establishment, multi-platform feature fusion and downstream prediction.
The overall process can be described as follows, divided into four steps:

Firstly, a lightweight backbone, ResNet-18 \cite{he2016deep}, is adopted as the feature extractor. Each platform inputs observations from various angles of the identical scene into the backbone,  generating features with generally similar content but localized differences in detail.

Secondly, leveraging the features generated by the fourth layer of the feature extractor, the SMIM module calculates a self-information confidence score locally to determine the collaboration request. Then, based on the feature relationships between different platforms, it interactively generates mutual-information match scores for the selection of collaborative platforms.

Thirdly, the RFF module captures the correlation between local features and those from the selected collaborative platform. It filters the locally required features and facilitates the fusion of misaligned features.

Finally, the fused features are fed into the downstream decoder, generating predicted results based on collaborative perception.

\subsection{Self-Mutual Information Match Module}

\begin{figure}
	\centering
	\includegraphics[width=0.5\textwidth]{"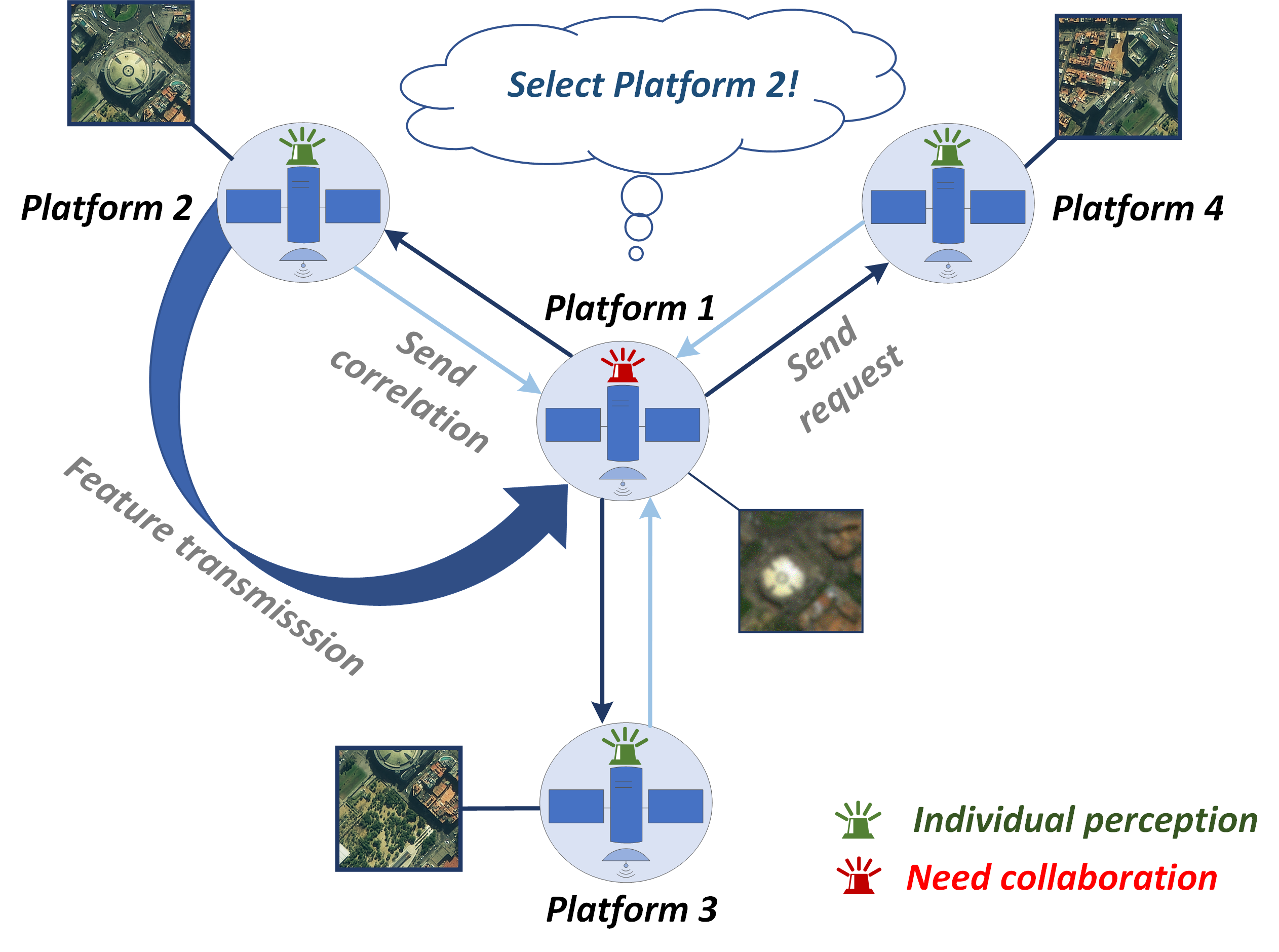"}
	\caption{\textbf{The schematic diagram of the SMIM module's function.} Each remote sensing platform in the collaborative network independently makes autonomous decisions based on its local observations. For instance, platform 1 seeks collaboration by sending requests to other platforms and selects a suitable supporter based on the correlation received in response. Conversely, platforms 2, 3, and 4 carry out the perception task individually. }
	\label{figure.3}
\end{figure}

\begin{figure*}
	\centering
	\includegraphics[width=0.9\textwidth]{"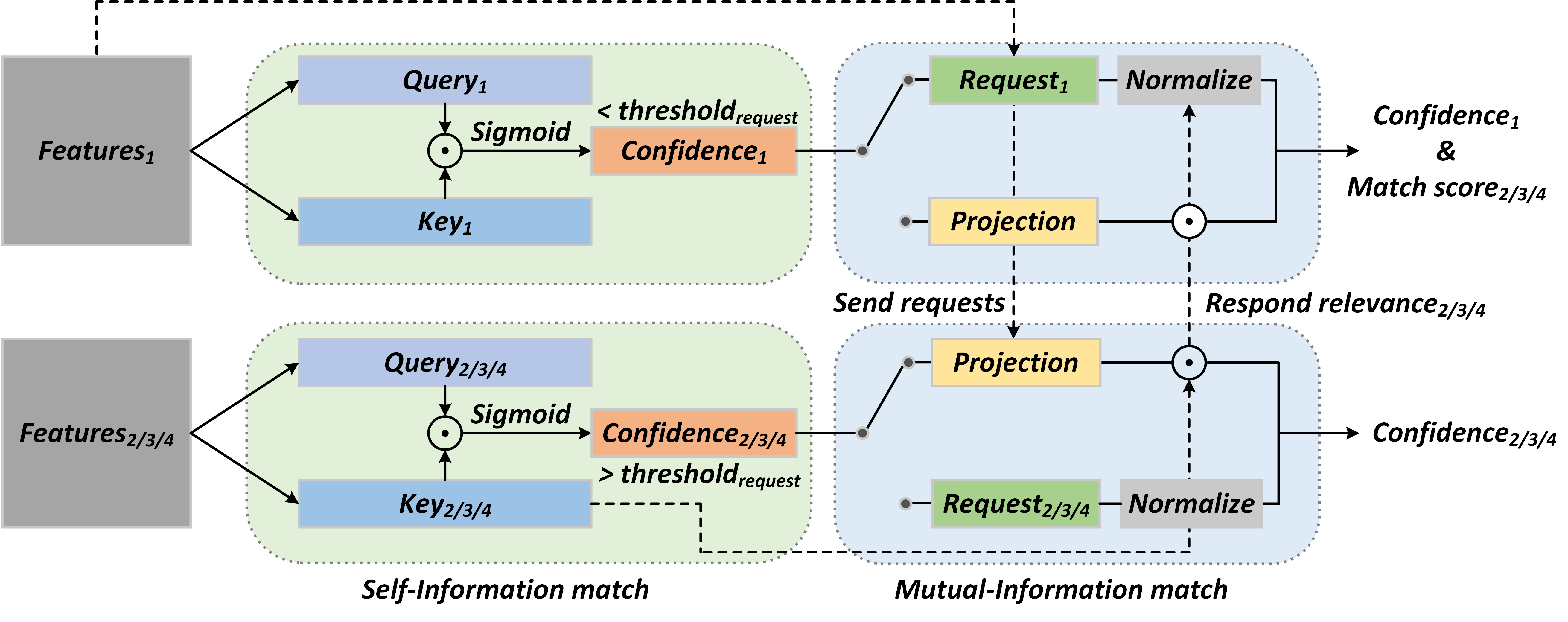"}
	\caption{The process framework of SMIM module. The module is divided into two stages: self-information match, depicted in the left part, and mutual-information match, depicted in the right part. }
	\label{figure.4}
\end{figure*}

A reliable information collaboration strategy should be able to ascertain the necessity for request and select appropriate collaborative platforms, precisely as the description in Fig.\ref{figure.3}. Additionally, the interaction cost must be taken into account. To address these concerns, we design a SMIM module that evaluates the need for collaboration based on the sufficiency of local information and identifies suitable supporters based on the relevance of supplementary features from the candidates. This guarantees the optimization of perceptual gains while also reducing communication expenditures.
The components of the SMIM module are depicted in Fig.\ref{figure.4}.  

Initially, platform $i$ encodes local features $x_i$ into query $q_i$ and key $k_i$ vector as the inputs of the self-information match stage:
\begin{equation}
	\begin{aligned}
		& q_i=Q\left(x_i ; \theta_q\right) \in \mathbb{R}^q ; i \in 1, \ldots, N; \\
		& k_i=K\left(x_i ; \theta_k\right) \in \mathbb{R}^k ; i \in 1, \ldots, N,
	\end{aligned}
\end{equation}
where $Q$ and $K$ refer to the query and key encoders, parameterized by $\theta_q$ and $\theta_k$, respectively. Notably, the collaborative group comprises a total of N platforms. Additionally, the vector space of query and key is represented by $\mathbb{R}$.

Subsequently, the dot-product self-attention mechanism, denoted by $\varphi$, is utilized to calculate the correlation $c_i$ between $q_i$ and $k_i$. This correlation indicates the amount of local information for agent $i$.

Then, through the sigmoid activation, the correlation is compressed into the interval $\left[0, 1\right]$ to generate the self-information confidence score $p_i$. This score represents the probability that the platform does not require collaboration:

\begin{equation}
	\begin{aligned}
		& \text{$p_i$} =\mathit{Sigmoid}\left(c_i\right)=\mathit{Sigmoid}\left(q_i^T k_i\right) \\
		& =\frac{1}{1+e^{-q_i^T k_i}} ; i \in 1, \ldots, N.
	\end{aligned}
\end{equation}

During the inference phase, a predefined request threshold is set to determine whether collaboration is necessary. If $p_i$ exceeds the threshold, it indicates that the local information is sufficient for the perceptual task independently without collaboration. This approach conserves communication resources by avoiding unnecessary requests for subsequent interactions.

\begin{equation}
	\begin{aligned}
		& \text{$p_i$} =\mathit{Sigmoid}\left(c_i\right)=\mathit{Sigmoid}\left(q_i^T k_i\right) \\
		& =\frac{1}{1+e^{-q_i^T k_i}} ; i \in 1, \ldots, N.
	\end{aligned}
\end{equation}

When platform $i$ seeks assistance, it generates a compact request vector $r_i$ by condensing the required features into a lower dimension $r$, which is significantly smaller than $k$, the feature dimension of the key generated in the self-information match stage:
\begin{equation}
	r_i=R\left(x_i ; \theta_r\right) \in \mathbb{R}^r ; i \in 1, \ldots, N,
\end{equation}
where $R$ denotes the request encoders, parameterized by $\theta_r$. This compression results in a significant reduction in transmission cost. Then platform i broadcasts its request to the candidates for collaboration.

In the mutual-information match stage, the SMIM module calculates the relevance between each candidate and requester to select the appropriate supporter. By re-utilizing the key $k_j$ generated in the self-information match stage, candidate $j$ can directly respond to the request $r_i$ from the requester, platform $i$, without incurring repeated feature encoding.
To handle inconsistent vector dimensions, the request $r_i$ is first projected onto the feature space with the same dimension as the key $k_j$ using the projection matrix $W_\alpha$. After the dot product operation, each candidate then feeds back the relevance to the requester, which is subsequently normalized to obtain the mutual-information match score $s_{ij}$.
Intuitively, a higher score $s_{ij}$ indicates that the candidate (platform $j$) can provide more informative information to the requester (platform $i$).

\begin{equation}
	s_{i j}=\mathit{Softmax} \left(r_i^T W_\alpha k_j\right)=\frac{e^{r_i^T W_\alpha k_j}}{\sum_{j=1, j \neq i}^N e^{r_i^T W_\alpha k_j}}
\end{equation}

In order to prioritize more helpful platforms for the requester and avoid low-yield interactions, a collaboration threshold is set to $\frac{1}{N-1}$. During the inference stage, the candidates whose mutual-information match scores exceed the collaboration threshold are designated as supporters and subsequently engaged in perceptive interaction with the requester.

\subsection{Related Feature Fusion Module}
\begin{figure*}
	\centering
	\includegraphics[width=1.0\textwidth]{"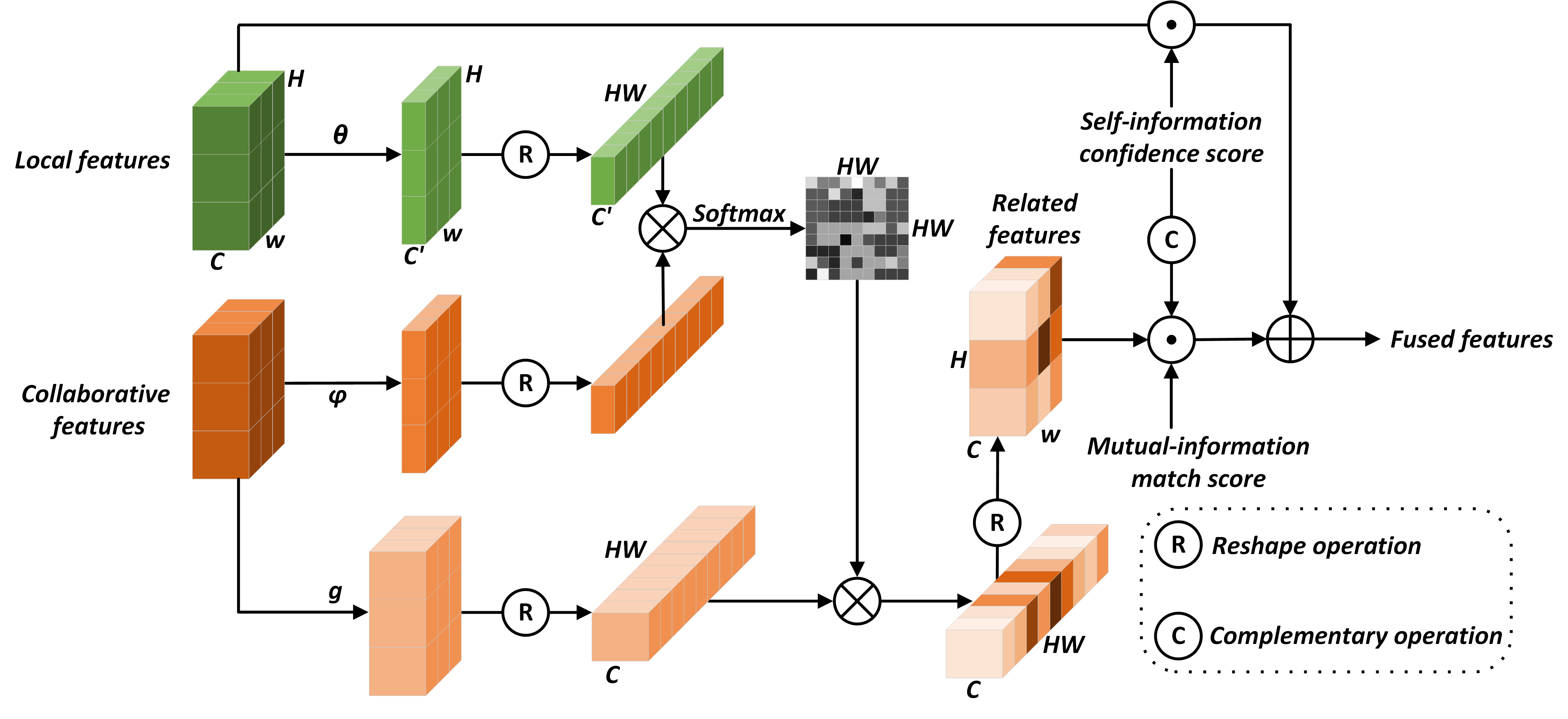"}
	\caption{The exhibition of RFF module. The reshape operation converts the image features into a sequence of features. In addition, the complementary operation can be simply regarded as subtracting the input from 1. }
	\label{figure.5}
\end{figure*}

Due to discrepancies in observation scope and perspective among various platforms, there exists the problem of inconsistent feature representation within the same scene. 
The naive concatenation or addition of feature fusion can contribute to feature misalignment. This misalignment disrupts the original observation information and ultimately triggers a decrease in perception accuracy. To this end, inspired by Non-local Neural Networks \cite{wang2018non}, our paper proposes a RFF module that utilizes the requester's features as a reference to select the correlated portion of the collaborative feature for fusion. By utilizing both the requester's and supporter's features, along with the self-information confidence score and mutual-information match score as inputs in this module, inter-dependencies between the elements of both feature sequences are captured for more effective and accurate feature fusion. The implementation details for the RRF module mentioned above are exhibited in Fig.\ref{figure.5}.

The generic calculation of related features is listed as follows:
\begin{equation}
	\begin{aligned}
		& F^r=\frac{1}{N(F)} \sum h\left(F^l, F^c\right) g\left(F^c\right) \\
		& =\frac{1}{N(F)} \theta\left(F^l\right) \varphi\left(F^c\right)^T g\left(F^c\right),
	\end{aligned}
\end{equation}
where $F^l$, $F^c$ and $F^r$ represent local features, collaborative features and related features, respectively. The pairwise function $h$ computes the relationship between $F^l$ and $F^c$, and the function  $N(F)$ normalizes the relationship to obtain the affinity matrix. Ultimately, the related feature $F^r$ is generated by the calculated affinity matrix and collaborative feature $F^c$. As to the concrete operation, the representation of input features is computed using linear embedding functions $\theta$, $\varphi$, and $g$, implemented as $1\times1$ convolution. Moreover, $h$ is implemented as a dot product manner and softmax is selected as the normalization function.

The specific steps of conducting the RFF module are demonstrated below.

Firstly, the local features $F^l$ and collaborative features $F^c$ are embedded by their respective encoders $\theta$ and $\varphi$, projecting the channel-wise dimension from $C$ to $C^{\prime}$, $C^{\prime}<< C$. This reduction in dimension reduces the complexity of subsequent affinity matrix computation. Additionally, the collaborative features are embedded by encoder $g$ and fixed in the original dimension for subsequent feature selection.
\begin{equation}
	\begin{aligned}
		& \theta\left(F^l\right), \varphi\left(F^c\right) \in \mathbb{R}^{H \times W \times C^{\prime}},
		g\left(F^c\right) \in \mathbb{R}^{H \times W \times C}
	\end{aligned}
\end{equation}

Secondly, a global cross-attention operation is performed on the flattened embeddings $\theta(F^l)$, $\varphi(F^c)$, calculating the relationship between each feature vector of the supporter and requester, and achieving fine-grained feature match. The resulting relationship is then normalized using softmax to obtain the affinity matrix $A \in \mathbb{R}^{H W \times H W}$, which represents the relationship between the entire feature sequences.
\begin{equation}
	\begin{aligned}
		& A=\mathit{Softmax} \left(\theta\left(F^l\right)^T \varphi\left(F^c\right)\right)=\left[A_{i, j}\right]_{H W^* H W} \\
		& A_{i, j}=\frac{\theta\left(F^l\right)_i{ }^T \varphi\left(F^c\right)_j}{\sum_{j \in H W} \theta\left(F^l\right)_i^T \varphi\left(F^c\right)_j} \in \mathbb{R}^{1 \times H W}
	\end{aligned}
\end{equation}

Thirdly, the related feature $F^r=A \times g\left(F^C\right) \in \mathbb{R}^{H \times W \times C}$ is obtained with the affinity matrix and embedded collaborative features.

Ultimately, the fused features $O^f$ are obtained by integrating local features with related features, along with the self-information confidence score and mutual-information match score determined by the SMIM module. This fused feature is then inputted into the downstream decoder to make predictions.
\begin{equation}
	O_i=p_i \cdot F_i^l+\left(1-p_i\right) \cdot \mathit{request}_i \cdot \sum_{j=1, j \neq i}^n s_{i j} \cdot F_j^r
\end{equation}
\subsection{Training Strategy}
Throughout the training process, it is necessary to instruct the platform on identifying the collaboration opportunities and selecting the appropriate supporter, which is a dynamic decision problem. Reinforcement learning is a widely employed approach to tackle such dynamic strategies \cite{mou2021deep, feng2021deep}. However, its effectiveness heavily depends on the reward mechanism design, presenting a significant challenge in collaborative perception. In contrast to intricate reinforcement learning methods, our proposed DCP-Net optimizes collaborative strategies solely based on the supervision of the ground truth from the downstream task.   

To implement the collaboration strategy, DCP-Net utilizes centralized training and distributed inference. During the training process, DCP-Net combines the features from all platforms and quantitatively evaluates the effect of each candidate's features on perception enhancement. In the inference stage, the principles designed in the SMIM module dynamically filter out unnecessary collaboration opportunities and candidates.
In this way, the requester learns to make decisions that maximize downstream prediction improvements by assessing local observations and interacting with the candidates.

Additionally, our experiments choose semantic segmentation as the downstream task and utilize cross-entropy loss for optimization. Throughout the iterations of loss minimization, DCP-Net continuously promotes its ability to select information and fuse features, ultimately achieving peak performance in collaborative perception. The overall loss function in our collaborative perception network during the training process is:
\begin{equation}
	\begin{aligned}
		\mathit{Loss} &= L(y_i, \hat{y_i}) = L(\mathit{decoder}(O_i), \hat{y_i}) \\
		&= L(\mathit{decoder}(p_i \cdot F_i^l + (1-p_i) \sum_{j=1, j \neq i}^n {s_{ij}} \cdot F_j^r), \hat{y_i}).
	\end{aligned}
\end{equation}

The analysis below delves into the reasons for DCP-Net's ability to achieve interactive strategy training solely through downstream supervision. The objective function optimization aims to minimize the loss in the downstream task, which heavily relies on the quality of the fused feature provided to the decoder. In other words, minimizing the loss corresponds to obtaining the optimal fused features while enhancing the feature quality primarily origins from integrating collaborative features during collaborative perception. The ideal fused features should follow a strategy that promotes perception while minimizing redundancy and interference. Consequently, in the SMIM module, when additional information is required to enhance perception, the calculated self-information confidence score is significantly lower, and the mutual-information match score assigned to candidates with greater perception improvement increases substantially. Additionally, the RFF module extracts more relevant and helpful information from collaborative features to achieve better feature fusion.

It is worth noting that the collaborative strategy supervised solely based on the ground truth of the downstream task provides convenience in terms of dataset availability. This approach eliminates the need for human intervention in identifying optimal collaboration opportunities and partners, thus reducing the challenge of sample annotation.

\section{Experiments}
This section is organized into four parts. Firstly, the methodology for constructing datasets for multi-platform collaborative perception is presented. Secondly, an overview of the experimental environment is provided. Thirdly, the data evaluation method employed in this article is explained. Lastly, the pertinent experimental details are outlined.
\subsection {The Construction of Datasets}
We evaluate DCP-Net through collaborative semantic segmentation tasks on simulated multi-platform datasets for aerial and satellite remote sensing, respectively, as well as a real joint observation dataset, DFC23 \cite{dfc2023}.

\begin{figure}[htbp]
	\centering
	\includegraphics[width=0.45\textwidth]{"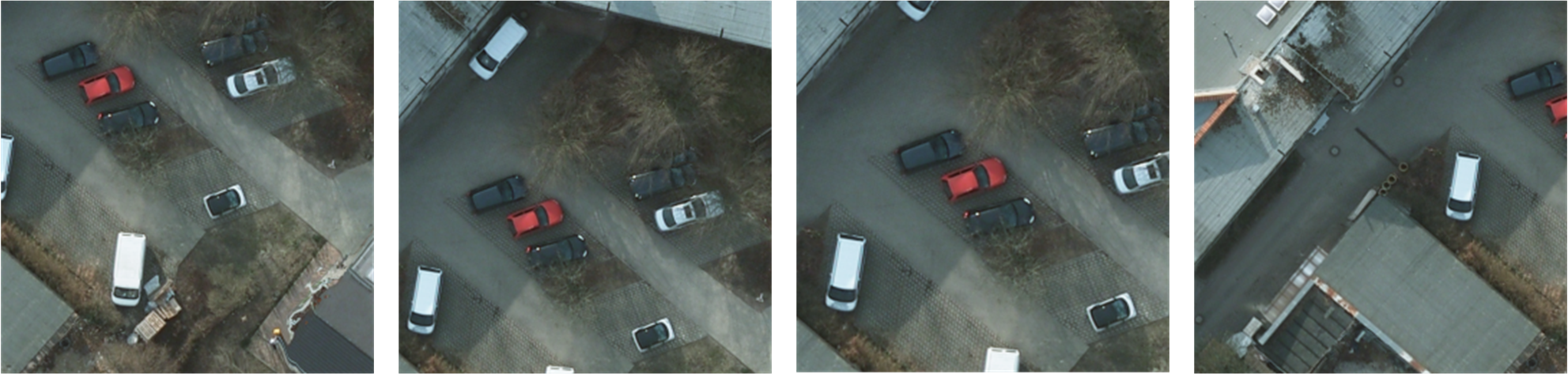"}
	\caption{Simulated multi-platform observation dataset of ISPRS Potsdam.}
	\label{figure.6}
\end{figure}

\begin{figure}[htbp]
	\centering
	\includegraphics[width=0.45\textwidth]{"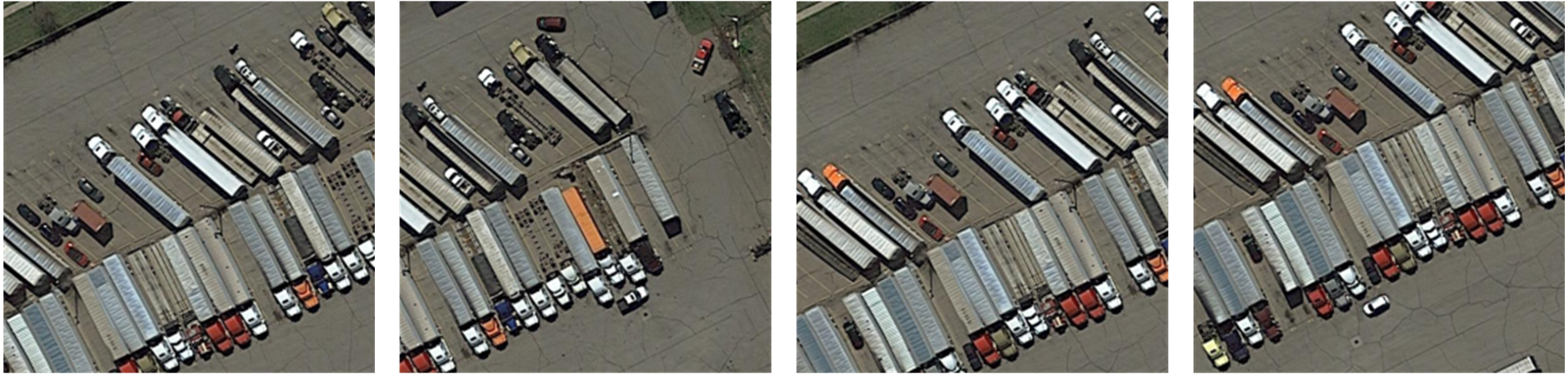"}
	\caption{Simulated multi-platform observation dataset of iSAID.}
	\label{figure.7}
\end{figure}

Due to the current scarcity of datasets for multi-platform collaborative observations, we generate analogous multi-platform datasets by utilizing existing datasets, including ISPRS Potsdam \cite{rottensteiner2014isprs} and iSAID \cite{waqas2019isaid}. This process is accomplished by following the steps outlined below:

Firstly, a full-size remote sensing image is cropped into multiple subsets of size 1024$\times$1024 by sliding windows. Secondly, four observation views of size 512$\times$512 are randomly cropped from each 1024$\times$1024 subset. By implementing these steps, a collection of images with overlapped observation ranges of identical scenes are obtained, forming the simulated datasets. The Fig.\ref{figure.6}, \ref{figure.7} below visually illustrate the resulting simulated datasets.
\begin{figure}[htbp]
	\centering
	\includegraphics[width=0.45\textwidth]{"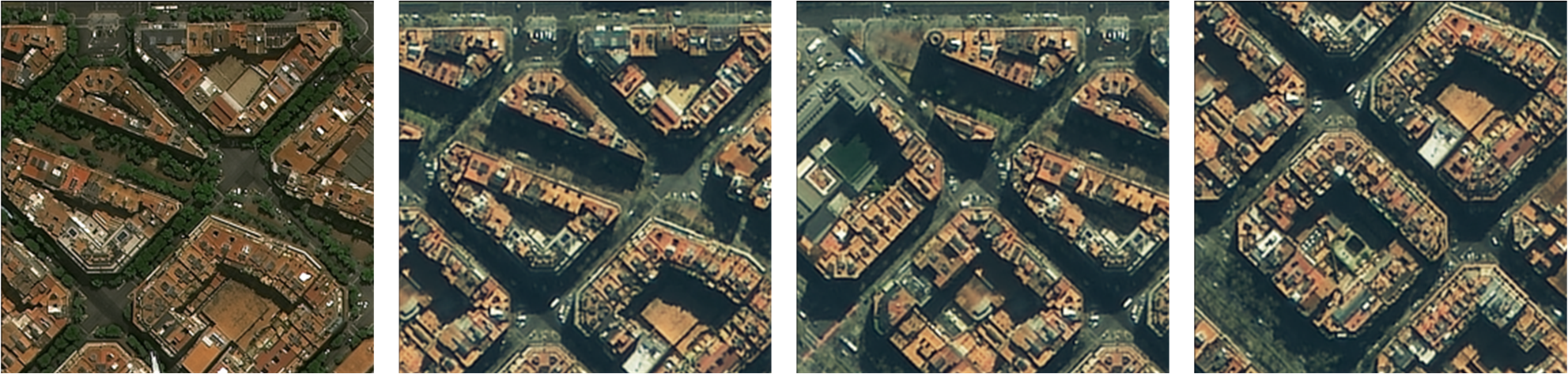"}
	\caption{Multi-platform joint observation dataset of DFC23.}
	\label{figure.9}
\end{figure}

\begin{figure*}
	\centering
	\includegraphics[width=1.0\textwidth]{"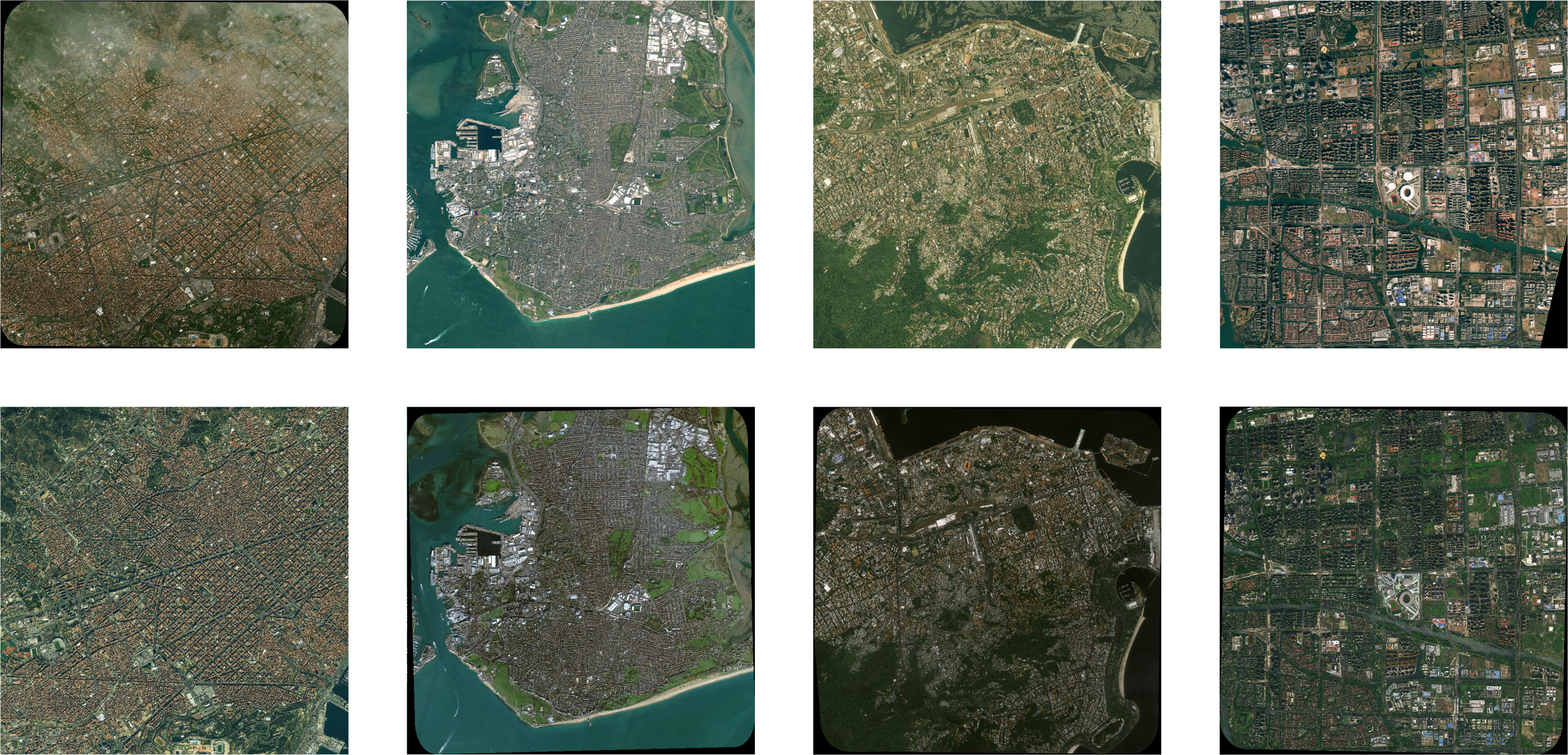"}
	\caption{The images taken by SuperView-1 and Gaofen-2 are presented in the row 1 and row 2, respectively.}
	\label{figure.8}
\end{figure*}

The real multi-platform observation dataset is created with the help of DFC23, jointly collected by the SuperView-1 and Gaofen-2 satellites. The dataset primarily comprises urban building clusters for collaborative semantic segmentation \cite{Huang_2022_CVPR}, aimed at assessing the practical effectiveness of various methods.
The Fig.\ref{figure.8} depicts the full-scale observation captured by SuperView-1 and Gaofen-2. Notably, differences exist in observation perspective, imaging characteristics, and resolution, making the dataset highly challenging and reflective of real-world scenarios.

As displayed in Fig.\ref{figure.9}, column 1 presents the image captured by SuperView-1, while columns 2 to 4 showcase images captured by Gaofen-2. The image in column 2, taken at the same geographic coordinate as column 1, exhibits significant differences in observation perspective, observation range and imaging payload. The nearby captured images, listed in columns 3 and 4, serve as similar interferences to verify DCP-Net's ability to filter out unnecessary additional information.

We have examined three experimental modes, categorized as homogeneous complete information supplement (Homo-CIS), homogeneous partial information supplement (Homo-PIS) and heterogeneous partial information supplement (Hetero-PIS). Details for each mode are outlined below.

\begin{figure}[t!]
	\centering
	\begin{minipage}[b]{0.45\textwidth}
		\centering
		\includegraphics[width=1.0\textwidth]{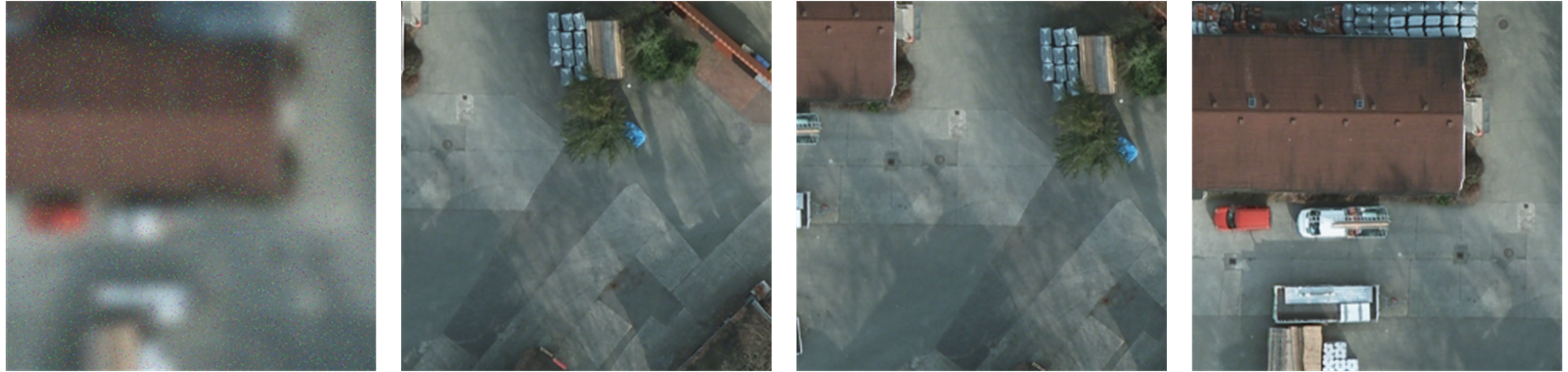}
		\caption{Visualization of the Homo-CIS mode.}
		\label{figure.10}
	\end{minipage}
	\begin{minipage}[b]{0.45\textwidth}
		\centering
		\includegraphics[width=1.0\textwidth]{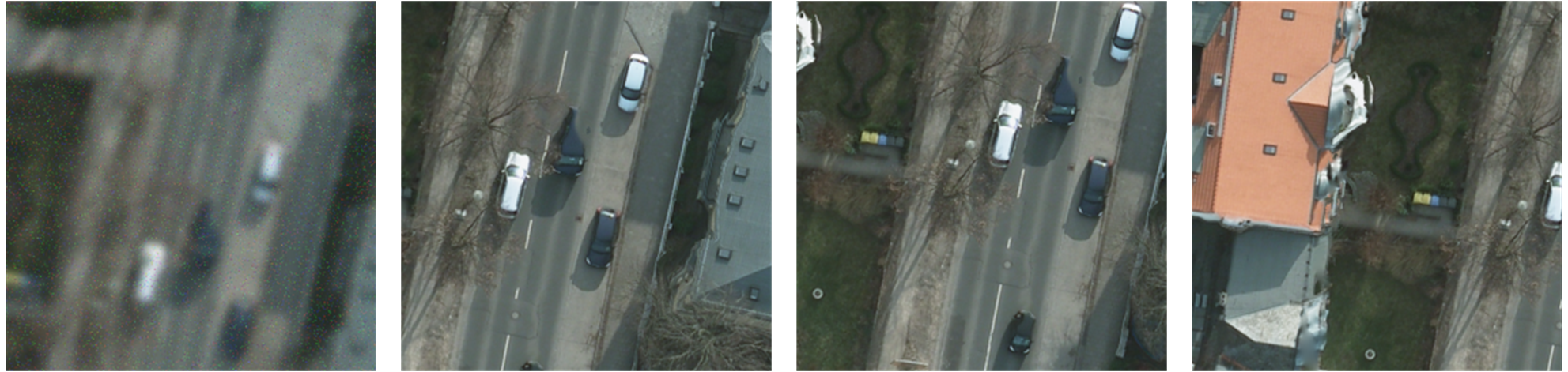}
		\caption{Visualization of the Homo-PIS mode.}
		\label{figure.11}
	\end{minipage}
	
	\begin{minipage}[b]{0.45\textwidth}
		\centering
		\includegraphics[width=1.0\textwidth]{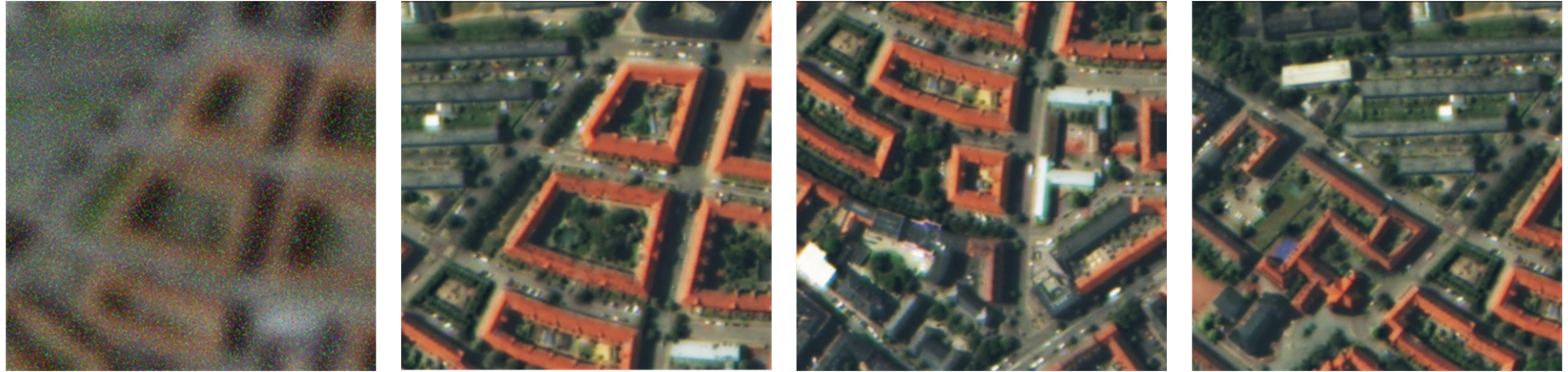}
		\caption{Visualization of the Hetero-PIS mode.}
		\label{figure.12}
	\end{minipage}
\end{figure}

The Homo-CIS mode aims to verify the accuracy of collaboration establishment and supporter selection. The term "homogeneous" implies that all platforms within the collaborative network possess the same imaging payload, while "complete" means that the requester can obtain all the necessary information comprehensively. In this mode, one platform suffers from imaging degradation, and its original, noise-free observation randomly appears among the other members within the group. Out of the four platforms, only one is selected as a potential victim of degradation. We introduce noise into 50\% of the images captured by this platform and randomly replace another platform's observations with the non-degraded original images. As depicted in Fig.\ref{figure.10}, column 1 displays the degraded image captured by the selected platform. Columns 2 and 3 show the observation information from other platforms, while column 4 presents the original, noise-free observation image. Furthermore, it's worth noting that only the segmentation mask of the potentially degraded platform is used as supervision during the training process. An ideal goal of the Homo-CIS mode is that the platform is capable of recognizing the degradation of local observations and utilizing the original information provided by other platforms to compensate for its perception. 

In the Homo-PIS mode, we eliminate the assumption that there exists original, noise-free observation of the degraded platform among the partners within the group. Due to differences in observation range among various remote sensing platforms, only the partial overlap of observations conforms to practical scenarios. In this mode, we investigate the improvement in perception with the help of supporters that only cover partially overlapped views. The degraded platform can only collaborate with candidates whose mutual-information match score exceeds the collaboration threshold. As depicted in Fig.\ref{figure.11}, the image in column 1 is marred by imaging noise, thereby leading to a degraded observation. In contrast, the images in the remaining columns offer clear and detailed representations of the overlapped scene. The ultimate goal of the Homo-PIS mode is to improve the perception of the degraded platform by effectively utilizing the partial-overlapped collaborative features available within the collaborative network.

Compared to the Homo-PIS mode, the Hetero-PIS mode is more complex, as it takes into account inconsistent imaging payloads among different platforms. The heterogeneity in payloads contributes to domain discrepancy, posing challenges in integrating diverse sources of information.
In this mode, we utilize observations from multiple satellites, each equipped with unique imaging capabilities, perspective biases, and range limitations. 
As shown in Fig.\ref{figure.12}, the images in columns 2-4 sourced from Gaofen-2 are used to collaborate with the degraded observations in column 1 sourced from SuperView-1.

\subsection{Implementation Details}
In our experimental setup, ResNet-18 is employed as the feature backbone for DCP-Net, alongside other baselines. For the segmentation task, FCN is served as the decoder. Additionally, the training and validation subsets of the datasets are divided as outlined in Table \ref{table.1}. In terms of optimization, the Adam optimizer is utilized with coefficient values of $\beta$ set to 0.9 and 0.999. The models undergo 50 epochs of training with a learning rate of $5 \times 10^{-5}$. All experiments are conducted on an RTX-3090 GPU with PyTorch version 1.7.1.
\begin{table}[]
	\setlength{\tabcolsep}{12pt}
	\centering
	\begin{tabular}{llll}
		\hline
		dataset       & classes & train & val  \\ \hline
		Potsdam & 6       & 7200  & 2800 \\ \hline
		iSAID         & 16      & 19790 & 6289 \\ \hline
		DFC23          & 2       & 3688  & 1752 \\ \hline
	\end{tabular}
	\caption{Split three datasets into the train set and validation set.}
	\label{table.1}
\end{table}

\subsection{Evaluation Metrics}
Our study conducts evaluations of all models in the collaborative segmentation task, taking into account three key aspects: mean IoU, communication expense, and collaboration efficiency.

Mean Intersection over Union (mIoU). It is a widely used evaluation metric in image segmentation, quantifying the overlap between predicted and ground truth targets. IoU is computed by dividing the intersection area of the predicted and ground truth regions by their union, ranging between 0 and 1. MIoU represents the average IoU across multiple targets and offers a concise and comprehensive assessment of algorithm performance on diverse targets, facilitating comparisons between different algorithms. A higher mIoU indicates superior target localization and segmentation capabilities. 

Communication Cost (Comm. Cost). 
This metric considers the average amount of communication in the interactions of the collaborative perception task. Additionally, we employ an indicator called MBpf (MegaBytes per frame) to quantitatively measure the communication cost. The MBpf is determined by both the frequency of collaboration establishment and the number of transmitted features.

Collaboration Efficiency (CE). It is crucial to comprehensively consider the metrics mentioned above to strike a balance between perception enhancement and interaction cost in collaborative perception. Accordingly, the concept of CE is introduced as a measure of this balance:
\begin{equation}
	CE=\frac{\Delta mIoU}{Comm.\ Cost}.
\end{equation}
CE is defined as the ratio of the improvement in accuracy achieved in the collaborative semantic segmentation to the communication cost incurred during the interaction. Consequently, a higher value of $\Delta$mIoU and a lower Comm.\ Cost is desirable in collaborative perception, as it signifies a more effective and efficient collaboration.

\subsection{Collaborative Semantic Segmentation Experiments}
We quantitatively compare the experimental results of our proposal DCP-Net with several centralized and distributed baselines in three modes mentioned in Section \uppercase\expandafter{\romannumeral 4}.A.
The brief introduction of the baseline methods is as follows:

\textbf{No-Interaction} serves as a baseline for the independent execution of downstream tasks without any information interaction. 

\textbf{Concat-All} is a simple centralized model baseline that concatenates all extracted image features from different platforms for downstream tasks.

\textbf{Auxiliary-View Attention} employs an attention mechanism to fuse observations from all platforms with various weights.

\textbf{Random-Selection} randomly selects a platform to provide observation information as a perception helper. 

\textbf{Who2com} \cite{liu2020who2com} utilizes an attention mechanism to select a perception helper based on relevance each time. 

\textbf{When2com} \cite{liu2020when2com}, an extension of Who2com, is able to determine the suitable collaboration opportunity.
\subsubsection{Homogeneous Complete Information Supplementation}

In Homo-CIS mode, we investigate the improvement brought by the collaborative perception in the background of both the same imaging payload and the supplement of the corresponding complete original information. Table \ref{table.2} shows the performance of several baseline models and our proposed model.
In both simulated multi-platform Potsdam and iSAID datasets, compared to the No-Interaction baseline, all centralized methods improved the predicted mIoU. However, they require all observations as assistance each time, which leads to high bandwidth consumption. In terms of the distributed manner, these methods, except for Random-Selection, are able to approach, even surpass the performance of centralized methods with only 1/6 to 1/3 of the communication overhead of the centralized one. As to naive Random Selection, the platform establishes collaboration without any strategy and filters the best perception supporter, just like winning a lottery. Therefore, Random-Selection contributes to a slight improvement of prediction in the simulated multi-platform iSAID dataset but even behaves much worse than No-Interaction in the simulated multi-platform Potsdam dataset.

Among the listed methods, our proposed DCP-Net achieves the optimal mask prediction in the Homo-CIS mode. Our proposed DCP-Net improves the average mIoU by 14.71\% and 16.89\% in the multi-platform Potsdam dataset and the multi-platform iSAID dataset, respectively. Besides, DCP-Net consumes the least Comm. Cost during the process of collaborative perception in this mode. In other words, regarding the CE, DCP-Net behaves far ahead of other baselines, realizing the ideal goal of the most perception enhancement and the least transmission cost.

\begin{figure}
	\centering
	\includegraphics[width=0.5\textwidth]{"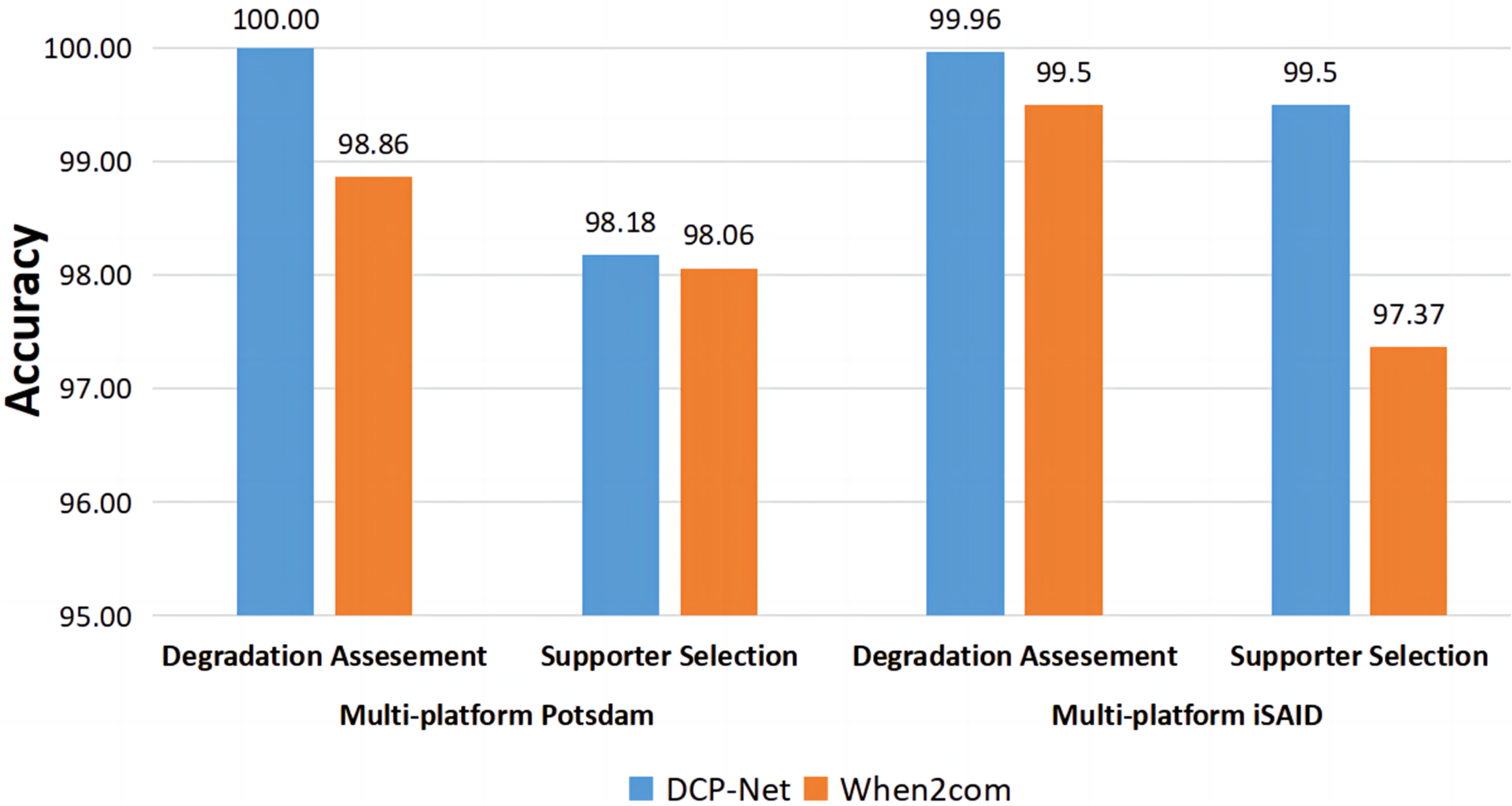"}
	\caption{The gaps of accuracy in assessing degradation and selecting the supporter between DCP-Net and When2com in the Homo-CIS mode.}
	\label{figure.13}
\end{figure}

The primary objective of the Homo-CIS mode is to validate our model's ability to identify collaboration opportunities and perception supporters. The other baselines, apart from Who2com and When2com, lack the capability of intelligent dynamic interaction. Furthermore, When2com, an upgraded version of Who2com, must establish continuous interaction among multiple platforms to determine collaboration opportunities and supporters simultaneously. We compare our model's accuracy in assessing collaborative opportunities and selecting supporters to When2com's. Fig.\ref{figure.13} illustrates that DCP-Net consistently outperforms When2com in evaluating collaboration opportunities and selecting supporters in the simulated multi-platform Potsdam and multi-platform iSAID datasets, thereby confirming the effectiveness of the SIMM module. In contrast to When2com, our DCP-Net assesses collaboration using a self-information confidence score and then employs mutual-information match scores to select supporters. This asynchronous operation sounds more reasonable and contributes to better performance while minimizing redundant communication waste. 

\begin{table*}[]
	\setlength{\tabcolsep}{4pt}
	\centering
	\begin{tabular}{
			>{\columncolor[HTML]{FFFFFF}}c 
			>{\columncolor[HTML]{FFFFFF}}c l
			>{\columncolor[HTML]{FFFFFF}}c 
			>{\columncolor[HTML]{FFFFFF}}c 
			>{\columncolor[HTML]{FFFFFF}}c 
			>{\columncolor[HTML]{FFFFFF}}c 
			>{\columncolor[HTML]{FFFFFF}}c l
			>{\columncolor[HTML]{FFFFFF}}c 
			>{\columncolor[HTML]{FFFFFF}}c 
			>{\columncolor[HTML]{FFFFFF}}c 
			>{\columncolor[HTML]{FFFFFF}}c 
			>{\columncolor[HTML]{FFFFFF}}c }
		\hline
		\multicolumn{2}{c}{\cellcolor[HTML]{FFFFFF}Homo-CIS    }                                                            &  & \multicolumn{5}{c}{\cellcolor[HTML]{FFFFFF}Postdam}                                                                                                                                                                                    &  & \multicolumn{5}{c}{\cellcolor[HTML]{FFFFFF}iSAID}                                                                                                                                                                                      \\ \hline
		\cellcolor[HTML]{FFFFFF}                                & \cellcolor[HTML]{FFFFFF}                                  &  & \multicolumn{3}{c}{\cellcolor[HTML]{FFFFFF}mIoU}                                                                & \cellcolor[HTML]{FFFFFF}                                     & \cellcolor[HTML]{FFFFFF}                              &  & \multicolumn{3}{c}{\cellcolor[HTML]{FFFFFF}mIoU}                                                                & \cellcolor[HTML]{FFFFFF}                                     & \cellcolor[HTML]{FFFFFF}                              \\ \hhline{~~~---~~~---~}
		\multirow{-2}{*}{\cellcolor[HTML]{FFFFFF}Type} & \multirow{-2}{*}{\cellcolor[HTML]{FFFFFF}Method} &  & \cellcolor[HTML]{FFFFFF}Noisy & \cellcolor[HTML]{FFFFFF}Normal & \cellcolor[HTML]{FFFFFF}Avg. & \multirow{-2}{*}{\cellcolor[HTML]{FFFFFF}Comm.\ Cost} & \multirow{-2}{*}{\cellcolor[HTML]{FFFFFF}CE} &  & \cellcolor[HTML]{FFFFFF}Noisy & \cellcolor[HTML]{FFFFFF}Normal & \cellcolor[HTML]{FFFFFF}Avg. & \multirow{-2}{*}{\cellcolor[HTML]{FFFFFF}Comm.\ Cost} & \multirow{-2}{*}{\cellcolor[HTML]{FFFFFF}CE} \\ \hline
		individual                                              & No-Interaction                                            &  & 50.18                                  & 65.09                                   & 57.38                                 & -                                                            & -                                                     &  & 38.77                                  & 49.33                                   & 44.24                                 & -                                                            & -                                                     \\ \hline
		\cellcolor[HTML]{FFFFFF}                                & Concat-All                                                &  & 55.48                                  & 65.92                                   & 60.60                                 & 1.500                                                        & 2.15                                                  &  & 44.57                                  & 52.20                                   & 48.47                                 & 1.500                                                        & 2.82                                                  \\
		\multirow{-2}{*}{\cellcolor[HTML]{FFFFFF}centralized}   & Auxiliary-View Attention                                  &  & 64.25                                  & 65.94                                   & 65.10                                 & 1.500                                                        & 5.15                                                  &  & 49.01                                  & \textbf{53.15}                          & 51.12                                 & 1.500                                                        & 4.59                                                  \\ \hline
		\cellcolor[HTML]{FFFFFF}                                & Random-Selection                                          &  & 49.58                                  & 60.67                                   & 54.98                                 & 0.500                                                        & -4.80                                                 &  & 40.03                                  & 51.06                                   & 45.73                                 & 0.500                                                        & 2.98                                                  \\
		\cellcolor[HTML]{FFFFFF}                                & Who2com                                                   &  & 64.59                                  & 65.90                                   & 65.25                                 & 0.500                                                        & 15.74                                                 &  & 43.66                                  & 50.06                                   & 46.91                                 & 0.500                                                        & 5.34                                                  \\
		\cellcolor[HTML]{FFFFFF}                                & When2com                                                  &  & 64.62                                  & 65.12                                   & 64.88                                 & 0.258                                                        & 29.13                                                 &  & 48.61                                  & 49.52                                   & 49.03                                 & 0.275                                                        & 17.42                                                 \\
		\multirow{-4}{*}{\cellcolor[HTML]{FFFFFF}distributed}   & Ours                                                      &  & \textbf{65.14}                         & \textbf{66.49}                          & \textbf{65.82}                        & \textbf{0.255}                                               & \textbf{33.10}                                        &  & \textbf{51.45}                         & 52.13                                   & \textbf{51.71}                        & \textbf{0.250}                                               & \textbf{29.88}                                        \\ \hline
	\end{tabular}
	\caption{Experimental results of baselines and DCP-Net in the Homo-CIS mode. Since there is no transmission cost in No-Interaction, the value of Comm. Cost and CE is represented as '$-$'. }
	\label{table.2}
\end{table*}

\subsubsection{Homogeneous Partial Information Supplementation}

The objective of the Homo-PIS mode is to verify the effectiveness of our proposed method in dealing with the challenge of misaligned collaborative features when there are only partially overlapped observations between platforms. However, due to the complexity of the overlapped collaborative features, accurately determining the optimal supporters becomes challenging. Therefore, the assessment of supporter selection accuracy is not conducted in this mode.

According to the results presented in Table \ref{table.3}, the margin of performance improvement for all methods is smaller with respect to No-Interaction, compared to more significant improvement observed in the Homo-CIS mode. The centralized methods outperform all other distributed methods, except for our DCP-Net in the multi-platform Potsdam and multi-platform iSAID datasets. Among the listed methods, our proposed DCP-Net achieves the optimal mask prediction and attains the highest collaboration efficiency in both datasets. This indicates that, unlike other distributed methods, DCP-Net still maintains a more remarkable perception improvement while consuming less communication cost, approximately 0.39MBpf. The experimental results quantitatively demonstrate that in the Homo-PIS mode, DCP-Net improves the average mIoU by 6.18\% and 7.34\% in the multi-platform Potsdam dataset and the multi-platform iSAID dataset, respectively.

It is worth noting that When2com incurs the lowest communication costs due to its inability to handle the misaligned collaborative features. Additionally, Table.\ref{table.3} demonstrates a substantial decrease in collaboration frequency in the multi-platform iSAID dataset and even negligible information exchange in the multi-platform Potsdam dataset, which is a concession of maintaining the original perception. Due to misalignment between collaborative  features and local features, a simple fusion can result in distorted observations. This distortion makes collaboration susceptible to being treated as interference in When2com. As to the other methods, they conduct collaborative perception in a compulsory manner and have adapted to utilizing misaligned collaborative features to enhance local perception. However, they are still inferior to our proposed DCP-Net, which can be attributed to the effectiveness of our designed RFF module.

\begin{table*}[]
	\setlength{\tabcolsep}{4pt}
	\centering
	\begin{tabular}{
			>{\columncolor[HTML]{FFFFFF}}c 
			>{\columncolor[HTML]{FFFFFF}}c l
			>{\columncolor[HTML]{FFFFFF}}c 
			>{\columncolor[HTML]{FFFFFF}}c 
			>{\columncolor[HTML]{FFFFFF}}c 
			>{\columncolor[HTML]{FFFFFF}}c 
			>{\columncolor[HTML]{FFFFFF}}c l
			>{\columncolor[HTML]{FFFFFF}}c 
			>{\columncolor[HTML]{FFFFFF}}c 
			>{\columncolor[HTML]{FFFFFF}}c 
			>{\columncolor[HTML]{FFFFFF}}c 
			>{\columncolor[HTML]{FFFFFF}}c }
		\hline
		\multicolumn{2}{c}{\cellcolor[HTML]{FFFFFF}Homo-PIS}                                                                &  & \multicolumn{5}{c}{\cellcolor[HTML]{FFFFFF}Postdam}                                                                                                                     &  & \multicolumn{5}{c}{\cellcolor[HTML]{FFFFFF}iSAID}                                                                                                                       \\ \hline
		\cellcolor[HTML]{FFFFFF}                                & \cellcolor[HTML]{FFFFFF}                                  &  & \multicolumn{3}{c}{\cellcolor[HTML]{FFFFFF}mIoU} & \cellcolor[HTML]{FFFFFF}                                     & \cellcolor[HTML]{FFFFFF}                              &  & \multicolumn{3}{c}{\cellcolor[HTML]{FFFFFF}mIoU} & \cellcolor[HTML]{FFFFFF}                                     & \cellcolor[HTML]{FFFFFF}                              \\  \hhline{~~~---~~~---~}
		\multirow{-2}{*}{\cellcolor[HTML]{FFFFFF}Type} & \multirow{-2}{*}{\cellcolor[HTML]{FFFFFF}Method} &  & Noisy    & Normal    & Avg.    & \multirow{-2}{*}{\cellcolor[HTML]{FFFFFF}Comm.\ Cost} & \multirow{-2}{*}{\cellcolor[HTML]{FFFFFF}CE} &  & Noisy    & Normal    & Avg.    & \multirow{-2}{*}{\cellcolor[HTML]{FFFFFF}Comm.\ Cost} & \multirow{-2}{*}{\cellcolor[HTML]{FFFFFF}CE} \\ \hline
		individual                                              & No-Interaction                                            &  & 48.47             & 63.30              & 55.37            & -                                                            & -                                                     &  & 38.59             & 50.42              & 44.69            & -                                                            & -                                                     \\ \hline
		\cellcolor[HTML]{FFFFFF}                                & Concat-All                                                &  & 50.56             & 64.18              & 56.92            & 1.500                                                        & 1.03                                                  &  & 41.83             & 51.61              & 46.89            & 1.500                                                        & 1.47                                                  \\
		\multirow{-2}{*}{\cellcolor[HTML]{FFFFFF}centralized}   & Auxiliary-View Attention                                  &  & 51.11             & 63.67              & 56.98            & 1.500                                                        & 1.07                                                  &  & 43.17             & \textbf{52.32}     & 47.91            & 1.500                                                        & 2.15                                                  \\ \hline
		\cellcolor[HTML]{FFFFFF}                                & Random-Selection                                          &  & 48.81             & 63.60              & 55.66            & 0.500                                                        & 0.58                                                  &  & 39.85             & 48.84              & 44.46            & 0.500                                                        & -0.46                                                 \\
		\cellcolor[HTML]{FFFFFF}                                & Who2com                                                   &  & 49.64             & 63.44              & 56.07            & 0.500                                                        & 1.40                                                  &  & 40.33             & 48.67              & 44.61            & 0.500                                                        & -0.16                                                 \\
		\cellcolor[HTML]{FFFFFF}                                & When2com                                                  &  & 49.71             & 61.77              & 55.33            & \textbf{0.015}                                               & -2.67                                                 &  & 36.42             & 48.43              & 42.59            & \textbf{0.110}                                               & -19.09                                                \\
		\multirow{-4}{*}{\cellcolor[HTML]{FFFFFF}distributed}   & Ours                                                      &  & \textbf{53.39}    & \textbf{64.84}     & \textbf{58.79}   & 0.388                                                        & \textbf{8.81}                                         &  & \textbf{45.56}    & 50.59              & \textbf{47.97}   & 0.385                                                        & \textbf{8.52}                                         \\ \hline
	\end{tabular}
	\caption{Experimental results of baselines and DCP-Net in the Homo-PIS mode.}
	\label{table.3}
\end{table*}

\subsubsection{Heterogeneous Partial Information Supplement Mode}
The objective of the Hetero-PIS mode is to test the performance of all methods in intricate real-world scenarios. The platforms in the provided DFC23 dataset exhibit variations in the observation angle, range, resolution, and imaging payload, making it more challenging than the Homo-PIS mode. In this mode, effective collaboration among heterogeneous data sources is crucial to compensate for local ambiguous observations of building clusters and improve prediction results.

As shown in Table \ref{table.4}, in comparison to the results of simulated experiments, the perception enhancement achieved by all centralized and distributed methods is constrained in such real-world situations. Due to the interference caused by the similar appearance of buildings in DFC23, Auxiliary-View Attention method, which is comparable to our proposed method in perceiving improvements in the Homo-CIS and Homo-PIS modes, faces a dilemma in collaboration and demonstrates inferior performance.
Nevertheless, our proposed method consistently outperforms other methods in the prediction accuracy with a relatively low Comm. Cost of 0.355 MBpf and undoubtedly owns the highest CE. These results demonstrate the generalizability and applicability of DCP-Net in real-world settings.

\begin{table}[!t]
	\setlength{\tabcolsep}{1pt}
	\centering
	\begin{tabular}{ccccccc}
		\hline
		\multicolumn{2}{c}{Hetero-PIS}                                               & \multicolumn{5}{c}{DFC23}                                                                      \\ \hline
		\multicolumn{1}{c}{\multirow{2}{*}{Type}} & \multirow{2}{*}{Method} & \multicolumn{3}{c}{mIoU}                 & \multirow{2}{*}{Comm.\ Cost} & \multirow{2}{*}{CE} \\ \cline{3-5}
		\multicolumn{1}{c}{}                               &                                  & Noisy & Normal & Avg.  &                                     &                              \\ \hline
		\multicolumn{1}{c}{individual}                     & No-Interaction                   & 54.94 & 61.04  & 57.88 & -                                   & -                            \\ \hline
		\multicolumn{1}{c}{\multirow{2}{*}{centralized}}   & Concat-All                       & 55.46 & 61.43  & 58.37 & 1.500                               & 0.33                         \\
		\multicolumn{1}{c}{}                               & Auxiliary-View Attention         & 55.50 & 62.22  & 58.85 & 1.500                               & 0.65                         \\ \hline
		\multicolumn{1}{c}{\multirow{4}{*}{distributed}}   & Random-Selection                 & 55.82 & 61.58  & 58.62 & 0.500                               & 1.48                         \\
		\multicolumn{1}{c}{}                               & Who2com                          & 54.28 & 61.63  & 57.91 & 0.500                               & 0.06                         \\
		\multicolumn{1}{c}{}                               & When2com                         & 55.74 & 61.82  & 58.77 & \textbf{0.265}                      & 3.36                        \\
		\multicolumn{1}{c}{}                               & Ours                             & \textbf{56.03} & \textbf{62.81}  & \textbf{59.39} & 0.355                               & \textbf{4.25}                \\ \hline
	\end{tabular}
	
	\caption{Experimental results of baselines and DCP-Net in the Hetero-PIS mode. }
	\label{table.4}
\end{table}

\subsection{Ablation Study}

\subsubsection{Designed Modules Ablation Analysis}

\begin{table*}[]
	\setlength{\tabcolsep}{6pt}
	\centering
	\begin{tabular}{cccccccccccc}
		\hline
		\rowcolor[HTML]{FFFFFF} 
		\multicolumn{12}{c}{\cellcolor[HTML]{FFFFFF}{\color[HTML]{333333} Homo-CIS}}                                                                                                                                                                                                                                                                                                                                                                                                  \\ \hline
		\rowcolor[HTML]{FFFFFF} 
		\multicolumn{2}{c}{\cellcolor[HTML]{FFFFFF} Components}                                                  & \multicolumn{5}{c}{\cellcolor[HTML]{FFFFFF} Postdam}                                                                                                                     & \multicolumn{5}{c}{\cellcolor[HTML]{FFFFFF} iSAID}                                                                                                                       \\ \hline
		\rowcolor[HTML]{FFFFFF} 
		\cellcolor[HTML]{FFFFFF}                                & \cellcolor[HTML]{FFFFFF}                               & \multicolumn{3}{c}{\cellcolor[HTML]{FFFFFF} mIoU} & \cellcolor[HTML]{FFFFFF}                                     & \cellcolor[HTML]{FFFFFF}                              & \multicolumn{3}{c}{\cellcolor[HTML]{FFFFFF} mIoU} & \cellcolor[HTML]{FFFFFF}                                     & \cellcolor[HTML]{FFFFFF}                              \\ \hhline{~~---~~---~}

		\rowcolor[HTML]{FFFFFF} 
		\multirow{-2}{*}{\cellcolor[HTML]{FFFFFF} SMIM} & \multirow{-2}{*}{\cellcolor[HTML]{FFFFFF} RFF} & Noisy    & Normal    & Avg.    & \multirow{-2}{*}{\cellcolor[HTML]{FFFFFF} Comm.\ Cost} & \multirow{-2}{*}{\cellcolor[HTML]{FFFFFF} CE} & Noisy    & Normal    & Avg.    & \multirow{-2}{*}{\cellcolor[HTML]{FFFFFF} Comm.\ Cost} & \multirow{-2}{*}{\cellcolor[HTML]{FFFFFF} CE} \\ \hline
		\rowcolor[HTML]{FFFFFF} 
		&                                                           & 50.18             & 65.09              & 57.38            & -                                                            & -                                                     & 38.77             & 49.33              & 44.24            & -                                                            & -                                                     \\
		\rowcolor[HTML]{FFFFFF} 
		& $\sqrt{}$                                                      & \textbf{65.49}    & \textbf{66.36}     & \textbf{65.92}   & 1.500                                                        & 5.69                                                  & \textbf{51.65}    & \textbf{52.96}     & \textbf{52.24}   & 1.500                                                        & 5.33                                                  \\
		\rowcolor[HTML]{FFFFFF} 
		$\sqrt{}$                                                       &                                                        & 64.74             & 65.87              & 65.31            & \textbf{0.255}                                               & 31.10                                                 & 50.11             & 51.37              & 50.69            & \textbf{0.248}                                               & 25.98                                                 \\
		\rowcolor[HTML]{FFFFFF} 
		$\sqrt{}$                                                       & $\sqrt{}$                                                      & 65.39             & \textbf{66.36}     & 65.87            & \textbf{0.255}                                               & \textbf{33.29}                                        & 51.45             & 52.13              & 51.71            & 0.250                                                        & \textbf{29.88}                                        \\ \hline
	\end{tabular}
	\caption{Ablation experiments on each designed module in the Homo-CIS mode.}
	\label{table.5}
\end{table*}

\begin{table*}[]
	\setlength{\tabcolsep}{6pt}
	\centering
	\begin{tabular}{cccccccccccc}
		\hline
		\rowcolor[HTML]{FFFFFF} 
		\multicolumn{12}{c}{\cellcolor[HTML]{FFFFFF}{\color[HTML]{333333} Homo-PIS}}                                                                                                                                                                                                                                                                                                                                                                                                \\ \hline
		\rowcolor[HTML]{FFFFFF} 
		\multicolumn{2}{c}{\cellcolor[HTML]{FFFFFF} Components}                                                  & \multicolumn{5}{c}{\cellcolor[HTML]{FFFFFF} Postdam}                                                                                                                     & \multicolumn{5}{c}{\cellcolor[HTML]{FFFFFF} iSAID}                                                                                                                       \\ \hline
		\rowcolor[HTML]{FFFFFF} 
		\cellcolor[HTML]{FFFFFF}                                & \cellcolor[HTML]{FFFFFF}                               & \multicolumn{3}{c}{\cellcolor[HTML]{FFFFFF} mIoU} & \cellcolor[HTML]{FFFFFF}                                     & \cellcolor[HTML]{FFFFFF}                              & \multicolumn{3}{c}{\cellcolor[HTML]{FFFFFF} mIoU} & \cellcolor[HTML]{FFFFFF}                                     & \cellcolor[HTML]{FFFFFF}                              \\  \hhline{~~---~~---~} \cline{8-10}
		\rowcolor[HTML]{FFFFFF} 
		\multirow{-2}{*}{\cellcolor[HTML]{FFFFFF} SMIM} & \multirow{-2}{*}{\cellcolor[HTML]{FFFFFF} RFF} & Noisy   & Normal   & Average   & \multirow{-2}{*}{\cellcolor[HTML]{FFFFFF} Comm.\ Cost} & \multirow{-2}{*}{\cellcolor[HTML]{FFFFFF} CE} & Noisy   & Normal   & Average   & \multirow{-2}{*}{\cellcolor[HTML]{FFFFFF} Comm.\ Cost} & \multirow{-2}{*}{\cellcolor[HTML]{FFFFFF} CE} \\ \hline
		\rowcolor[HTML]{FFFFFF} 
		&                                                        & 48.47            & 63.30             & 55.37              & -                                                            & -                                                     & 38.59            & 50.42             & 44.69              & -                                                            & -                                                     \\
		\rowcolor[HTML]{FFFFFF} 
		& $\sqrt{}$                                                      & \textbf{56.19}   & \textbf{64.54}    & \textbf{60.11}     & 1.500                                                        & 3.16                                                  & \textbf{47.20}   & 50.82             & \textbf{48.95}     & 1.500                                                        & 2.84                                                  \\
		\rowcolor[HTML]{FFFFFF} 
		$\sqrt{}$                                                       &                                                        & 49.98            & 63.48             & 56.19              & -                                              & -                                                     & 39.93            & 49.67             & 45.03              & \textbf{0.315}                                               & 1.08                                                  \\
		\rowcolor[HTML]{FFFFFF} 
		$\sqrt{}$                                                      &  $\sqrt{}$                                                      & 54.43            & 63.70             & 58.91              & \textbf{0.388}                                               & \textbf{9.12}                                         & 45.56            & \textbf{50.59}    & 47.97              & 0.385                                                        & \textbf{8.52}                                         \\ \hline
	\end{tabular}
	\caption{Ablation experiments on each designed module in the Homo-PIS mode.}
	\label{table.6}
\end{table*}

\begin{table}
	\setlength{\tabcolsep}{4pt}
	\centering
	\begin{tabular}{ccccccc}
		\hline
		\rowcolor[HTML]{FFFFFF} 
		\multicolumn{7}{c}{\cellcolor[HTML]{FFFFFF}{\color[HTML]{333333} Hetero-PIS}}                                                                                                                                                                                                                                                                                \\ \hline
		\rowcolor[HTML]{FFFFFF} 
		\multicolumn{2}{c}{\cellcolor[HTML]{FFFFFF} Components}                                                  & \multicolumn{5}{c}{\cellcolor[HTML]{FFFFFF} DFC23}                                                                                                                                                                          \\ \hline
		\rowcolor[HTML]{FFFFFF} 
		\cellcolor[HTML]{FFFFFF}                                & \cellcolor[HTML]{FFFFFF}                               & \multicolumn{3}{c}{\cellcolor[HTML]{FFFFFF} mIoU}                                                                   & \cellcolor[HTML]{FFFFFF}                                     & \cellcolor[HTML]{FFFFFF}                              \\ \hhline{~~|---|~~}
		\rowcolor[HTML]{FFFFFF} 
		\multirow{-2}{*}{\cellcolor[HTML]{FFFFFF} SMIM} & \multirow{-2}{*}{\cellcolor[HTML]{FFFFFF} RFF} & Noisy & Normal & Average & \multirow{-2}{*}{\cellcolor[HTML]{FFFFFF} Comm.\ Cost} & \multirow{-2}{*}{\cellcolor[HTML]{FFFFFF} CE} \\ \hline
		\rowcolor[HTML]{FFFFFF} 
		&                                                        & 54.94                                  & 61.04                                   & 57.88                                    & -                                                            & -                                                     \\
		\rowcolor[HTML]{FFFFFF} 
		& $\sqrt{}$                                                      & 55.96                                  & \textbf{63.29}                          & \textbf{59.66}                           & 1.500                                                        & 1.19                                                  \\
		\rowcolor[HTML]{FFFFFF} 
		$\sqrt{}$                                                       &                                                        & 55.73                                  & 62.21                                   & 58.89                                    & 0.360                                                         & 2.85                                                  \\
		\rowcolor[HTML]{FFFFFF} 
		$\sqrt{}$                                                       &  $\sqrt{}$                                                      & \textbf{56.03}                         & 62.81                                   & 59.39                                    & \textbf{0.355}                                               & \textbf{4.25}                                         \\ \hline
	\end{tabular}
	\caption{Ablation experiments on each designed module in the Hetero-PIS mode.}
	\label{table.7}
\end{table}

Extensive ablation experiments are conducted on three datasets to evaluate the effectiveness of our designed SMIM module and RFF module. 
As shown in Table \ref{table.5}, \ref{table.6}, \ref{table.7}, a consistent trend is observed across all three datasets. The DCP-Net containing only the SMIM module or the RFF module both demonstrates improved performance compared to the No-Interaction baseline. This provides preliminary evidence for the effectiveness of the respective modules. For a single-module DCP-Net, the model without the SMIM module losses the capability of dynamic interactions and conducts centralized collaborative perception, demonstrating the best performance. The centralized baselines perform much weaker than the DCP-Net without the SMIM module in perception improvement. This phenomenon reveals that our designed RFF module can leverage the collaborative features more efficiently. The DCP-Net without the RFF module saves a significant expense of communication overhead while still maintaining a relatively high level of predicted accuracy. The result exhibits the talent of the SMIM module in selecting the appropriate opportunity and perception supporter. However, in some complicated conditions, such as the multi-platform Potsdam dataset of the Homo-PIS mode, the DCP-Net without the RFF module exhibits weakness in misaligned feature fusion and gives up the opportunity for collaborative perception. The phenomenon also verifies the necessity of the RFF module. Furthermore, the model with two modules working simultaneously obtains the highest collaborative efficiency and achieves a win-win goal of improving performance with less information exchange. This further validates the effectiveness and excellent combined performance of the SMIM and RFF modules.

\subsubsection{Hyperparameter Ablation Analysis}
\begin{figure}[h]
	\centering
	\includegraphics[width=0.5\textwidth]{"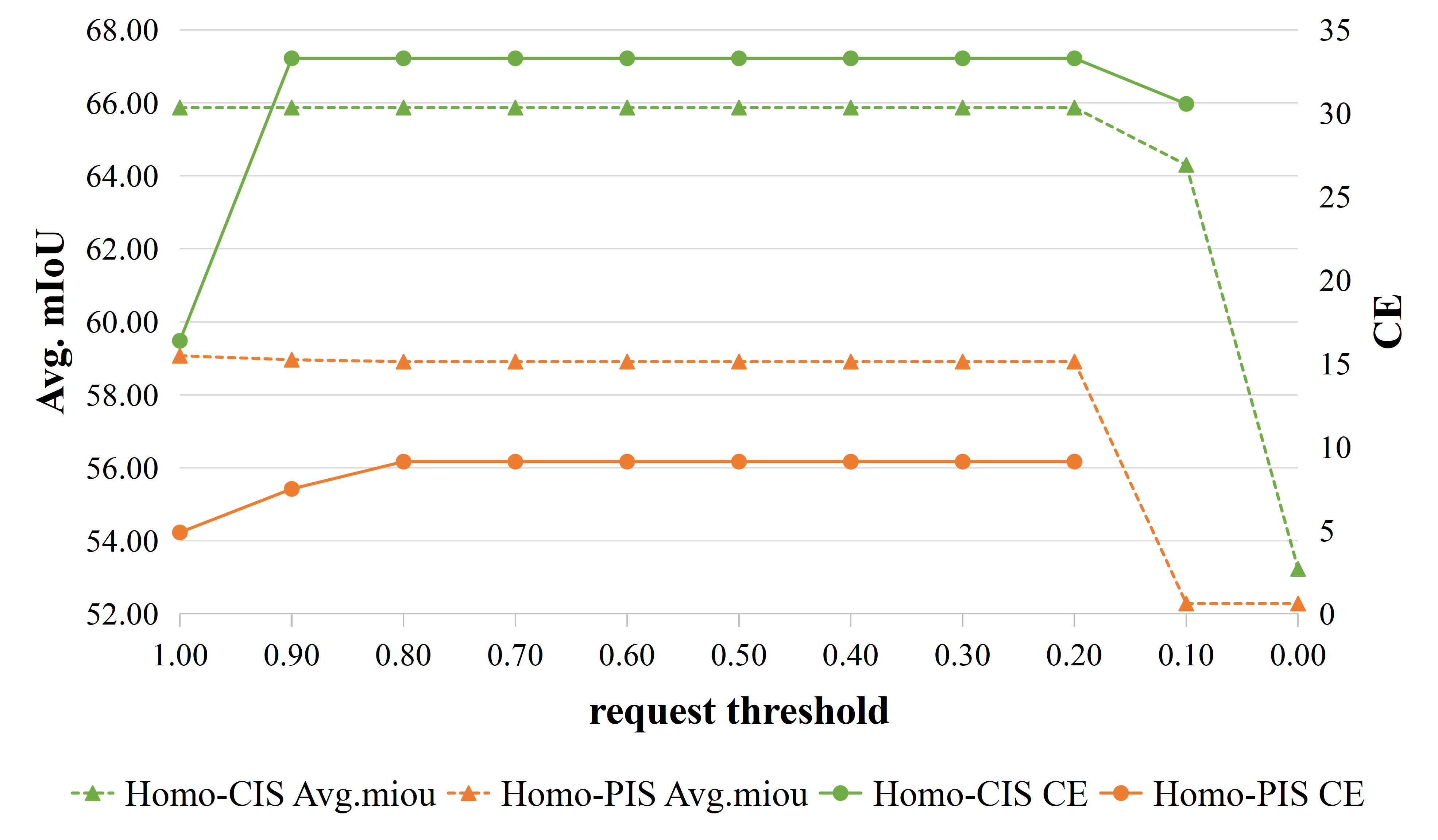"}
	\caption{Ablation experiments on the request threshold in the self-information match stage of the SMIM module.}
	\label{figure.14}
\end{figure}

In the SMIM module, the request threshold is predefined as a hyperparameter during the self-information match stage. The platform with a self-information confidence score below the request threshold is prompted to send collaboration requests to others. A proper threshold setting is crucial for establishing collaborative perception.
Therefore, an ablation experiment is performed to evaluate the impact of various thresholds on the performance of the multi-platform Potsdam dataset in the Homo-PIS mode, as shown in Fig.\ref{figure.14}.
When the request threshold is decreased from 1 to 0.9, there is a negligible decrease in the average mIoU, indicating that the performance remains relatively stable. However, there is a significant doubling in CE, suggesting that the system becomes more efficient in terms of collaboration.
Conversely, when the request threshold is changed from 0.2 to 0, both the average mIoU and CE experience a significant decline, which can be attributed to a sudden reduction in collaboration frequency.
Interestingly, a similar trend is observed in the Homo-PIS mode, confirming the findings in the Homo-CIS mode. To strike a balance between accuracy and efficiency, this paper selects a collaboration threshold of 0.8.

\begin{figure}[h]
	\centering
	\includegraphics[width=0.5\textwidth]{"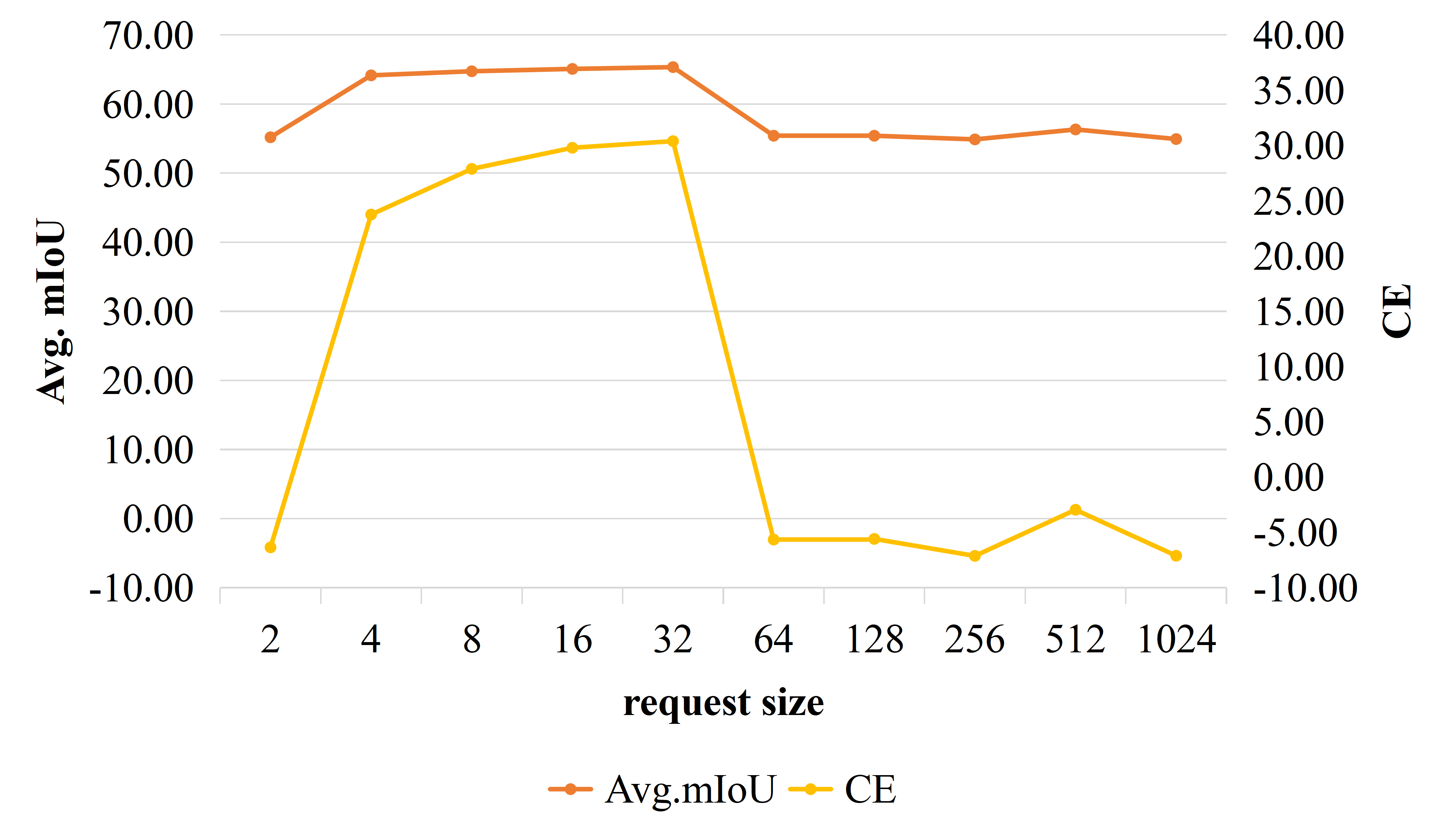"}
	\caption{Ablation experiments on the request size used in the mutual-information match stage of the SMIM module.}
	\label{figure.15}
\end{figure}

In the SMIM module, each mutual-information match score is calculated based on the compressed request and its corresponding key. The scale of compression may affect the later supporter selection. So it is essential to explore the influence of the request vector size on collaborative perception prediction. The ratio of compression can affect the subsequent supporter selection process. Hence, it is crucial to explore the influence of the request vector size on collaborative perception prediction.
To explore this, an ablation experiment is conducted using the multi-platform Potsdam dataset in the Homo-PIS mode, with the size of the request vector varied from 2 to 1024. Fig.\ref{figure.15} exhibits that when the request size is set to 32, there is a distinct turning point in the curve for both metrics. Our proposed DCP-Net achieves the best results in terms of semantic segmentation prediction and collaboration efficiency while maintaining a sub-minimal communication overhead. Consequently, the size of the compressed request is set at 32 in the SMIM module. 

\begin{figure*}[t!]
	\centering
	\begin{minipage}[b]{1\textwidth}
		\centering
		\includegraphics[width=1.0\textwidth]{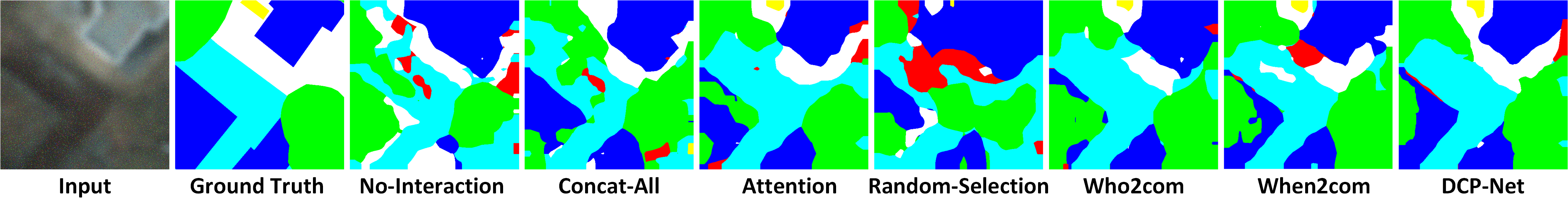}
		\caption{The visualization of results predicted by various baselines and DCP-Net in the Homo-CIS mode of the Potsdam dataset.}
		\label{subfig:potsdam-srms}
	\end{minipage}
	\begin{minipage}[b]{0.5\textwidth}
		\centering
		\includegraphics[width=\textwidth]{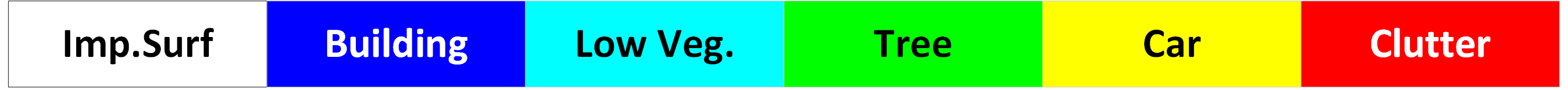}
		\label{subfig:seg-bar-potsdam}
	\end{minipage}
	
	\begin{minipage}[b]{1\textwidth}
		\centering
		\includegraphics[width=1.0\textwidth]{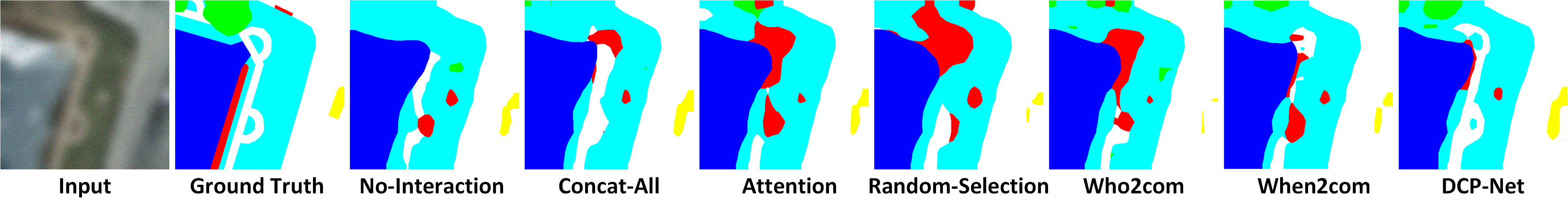}
		\caption{The visualization of results predicted by various baselines and DCP-Net in the Homo-PIS mode of the Potsdam dataset.}
		\label{subfig:potsdam-srmps}
	\end{minipage}
	
	\begin{minipage}[b]{1\textwidth}
		\centering
		\includegraphics[width=1.0\textwidth]{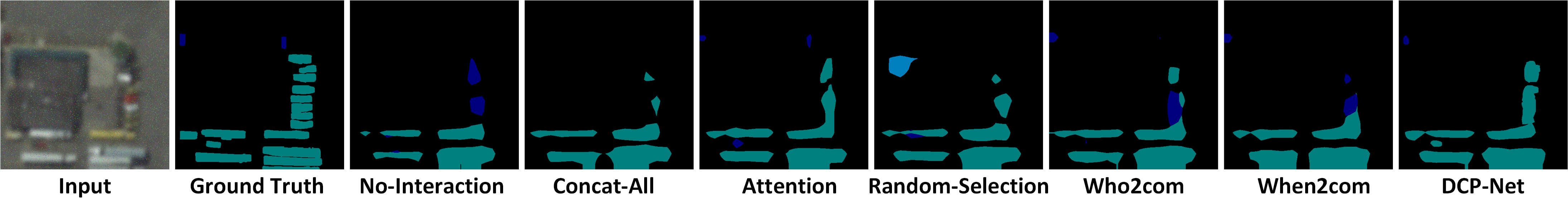}
		\caption{The visualization of results predicted by various baselines and DCP-Net in the Homo-CIS mode of the iSAID dataset.}
		\label{subfig:isaid-srms}
	\end{minipage}
	\begin{minipage}[b]{0.5\textwidth}
		\centering
		\includegraphics[width=\textwidth]{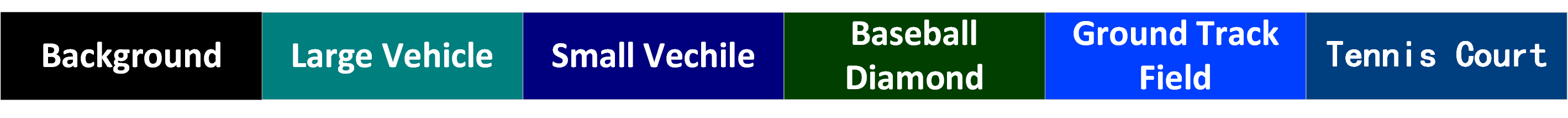}
		\label{subfig:seg-bar-isaid}
	\end{minipage}
	
	\begin{minipage}[b]{1\textwidth}
		\centering
		\includegraphics[width=1.0\textwidth]{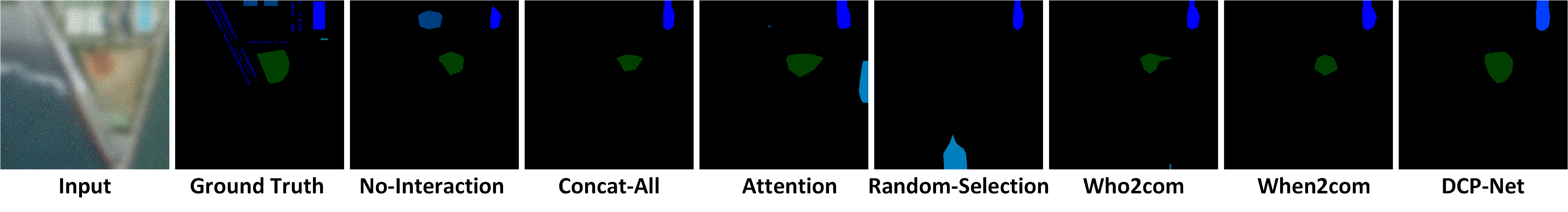}
		\caption{The visualization of results predicted by various baselines and DCP-Net in the Homo-PIS mode of the iSAID dataset.}
		\label{subfig:isaid-srmps}
	\end{minipage}
	
	\begin{minipage}[b]{1\textwidth}
		\centering
		\includegraphics[width=1.0\textwidth]{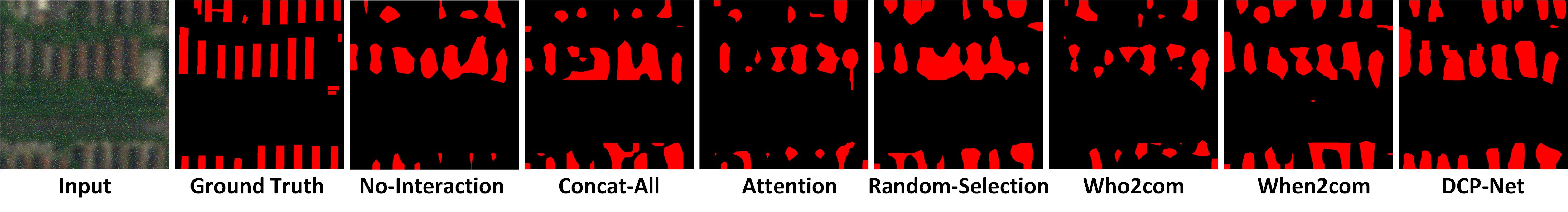}
		\caption{The visualization of results predicted by various baselines and DCP-Net in the Hetero-PIS mode of the DFC23 dataset.}
		\label{subfig:dfc-srmps}
	\end{minipage}
	\begin{minipage}[b]{0.2\textwidth}
		\centering
		\includegraphics[width=\textwidth]{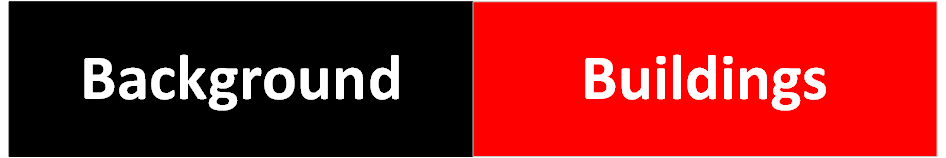}
		\label{subfig:seg-bar-dfc}
	\end{minipage}
	%
	\label{fig:results}
\end{figure*}

\subsection{Visualization}
The improvement of downstream predictions through the collaborative perception is presented in Fig.\ref{subfig:potsdam-srms},\ref{subfig:potsdam-srmps},\ref{subfig:isaid-srms},\ref{subfig:isaid-srmps},\ref{subfig:dfc-srmps} intuitively. 

In Fig.\ref{subfig:potsdam-srms},\ref{subfig:potsdam-srmps}, for the multi-platform Potsdam dataset of both modes, No-Interaction tends to classify uncertain objects resulting from image degradation as clutter and often misidentified blurry buildings as impervious surfaces. Additionally, it exhibits poor ability to distinguish between trees and low vegetation. Compared with other baselines, our DCP-Net can provide more accurate predictions for the foreground objects and boundary regions with the help of proper selection and fusion of collaborative features during the collaborative perception. 

In the multi-platform iSAID dataset of both modes, it is more challenging to discriminate 16 classes, particularly in a lower resolution, and we select only six classes for concise visualization. Due to the incomplete representation, No-Interaction mistakes some large vehicles for the small ones in Fig.\ref{subfig:isaid-srms} and directly ignores some tiny objects, such as the small vehicles in Fig.\ref{subfig:isaid-srmps}. In contrast with other methods, our DCP-Net is able to remedy the terrible condition mentioned above better and provide a more accurate regional prediction. 

In the practical scene of DFC23, 	
the buildings are densely clustered, and No-Interaction often misclassifies certain ranges of degraded observations as background indiscriminately. Furthermore, it is also tricky to improve the degraded perception with the help of heterogeneous collaborative features. Despite these challenging conditions, our DCP-Net still outperforms other methods. In Fig.\ref{subfig:dfc-srmps}, DCP-Net delivers a clearer demarcation between buildings and compensates for the predictions in the right bottom and the top. This visualization of predicted masks supports the effectiveness of DCP-Net in practical applications.

\section{Conclusion}
Motivated by the widespread deployment of intelligent remote sensing platforms, this paper proposes DCP-Net, a novel collaborative perception framework through feature interactions among multiple remote sensing platforms. DCP-Net leverages the designed SIMM module to establish adaptive collaboration with other platforms. In addition, the RFF module promotes the integration of features across multiple platforms to facilitate superior subsequent predictions.
Importantly, the entire process can be completed without the human intervention to determine the optimal collaboration opportunities and partners. Extensive experiments and visualization analyses are conducted on three datasets, namely Potsdam, iSAID and DFC23, which are redesigned in three difficulty modes. Comparative analysis with existing collaborative methods comprehensively demonstrates our proposed method's superiority.
%
\bibliographystyle{ieeetr}
\bibliography{references}

\end{document}